\documentclass[manuscript, screen]{jair}

\setcopyright{cc}
\acmDOI{10.1613/jair.1.xxxxx}

\JAIRAE{}
\JAIRTrack{}
\acmVolume{4}
\acmArticle{6}
\acmMonth{8}
\acmYear{2026}

\RequirePackage[
  datamodel=acmdatamodel,
  style=acmauthoryear,
  backend=biber,
  giveninits=true,
  uniquename=init
]{biblatex}

\usepackage{algorithm}
\usepackage{algpseudocode}
\usepackage{booktabs}
\usepackage{graphicx}
\usepackage{amsmath}

\usepackage{multirow}
\usepackage{xcolor}
\usepackage{listings}

\floatname{algorithm}{Algorithm}

\received{1 March 2026}
\received[accepted]{}

\begin{document}

\title[AutoWorldBuilder]{Fictional Worldbuilding: Multi-Agent LLM Collaboration with Hierarchical Context Compression and Iterative Review}

\author{Jingbo Chen}
\authornote{author1.}
\orcid{0009-0005-5177-5513}
\email{chenjingbo@nudt.edu.cn}
\affiliation{%
  \institution{National University of Defense Technology}
  \city{Nanjing}
  \state{Jiangsu}
  \country{China}
}

\author{He Wang}
\authornote{author2.}
\orcid{0009-0009-1274-8319}
\email{wanghe24@nudt.edu.cn}
\affiliation{%
  \institution{National University of Defense Technology}
  \city{Nanjing}
  \state{Jiangsu}
  \country{China}
}

\author{Wei Yuan}
\authornote{Corresponding Author.}
\orcid{0009-0004-2498-9447}
\email{yw5811827@126.com}
\affiliation{%
  \institution{National University of Defense Technology}
  \city{Nanjing}
  \state{Jiangsu}
  \country{China}
}

\author{Yuqiao Lai}
\authornote{author3.}
\orcid{0009-0006-9937-1232}
\email{laiyuqiao24@nudt.edu.cn}
\affiliation{%
  \institution{National University of Defense Technology}
  \city{Nanjing}
  \state{Jiangsu}
  \country{China}
}

\author{Zhenyan Lu}
\authornote{author4.}
\orcid{0009-0000-9126-7840}
\email{lzy\_25@nudt.edu.cn}
\affiliation{%
  \institution{National University of Defense Technology}
  \city{Nanjing}
  \state{Jiangsu}
  \country{China}
}

\renewcommand{\shortauthors}{Chen}

\begin{abstract}
Worldbuilding, the construction of coherent fictional worlds, is a foundational task in game design and literary creation. Large Language Models (LLMs) offer new possibilities for automated content generation, but their application to worldbuilding faces three challenges: context explosion that grows linearly with the building process, the tension between creative diversity and content consistency, and the absence of automated quality assurance.

This paper presents AutoWorldBuilder, a multi-agent collaborative system that addresses these challenges through five integrated components: a structured concept network with conflict detection; a DAG-based hybrid batch scheduler that groups tasks by semantic locality; a four-layer context compression mechanism achieving approximately 90\% token reduction; an iterative review system with specialized Auditor agents that improves proposal pass rates from 42\% to over 85\%; and a skill-driven agent architecture supporting zero-code extension with differentiated temperature configuration.

Two experiments across 20 diverse worldbuilding tasks, using GPT-OSS 120B and DeepSeek v3.2 as LLM backends, demonstrate a 95.0\% success rate. The system generated 56--103 self-consistent concepts per world in 18--31 minutes, with no conflicts detected by the review pipeline. The architectural patterns validated here, including layer-as-budget compression, semantic-locality scheduling, and separation of generation and review, may transfer to the broader class of knowledge-intensive, multi-agent LLM applications.
\end{abstract}

\maketitle

\section{Introduction}

\subsection{Background and Motivation}

Worldbuilding, the construction of coherent fictional worlds, is a foundational task in creative content production, including game design, literary creation, film and television production, and tabletop role-playing games (TRPGs). A complete, self-consistent fictional world typically encompasses multiple dimensions, including geography, intelligent races, magic or technology systems, social organizations, historical events, and cultural customs, and contains dozens to hundreds of interrelated concepts. For example, the Middle-earth of \textit{The Lord of the Rings} includes over 2,000 named locations and characters; the continent of Westeros in \textit{A Song of Ice and Fire} similarly features a complete system of geography, history, noble houses, and mythology. These worlds not only provide rich narrative backdrop but become a core source of a work's appeal.

However, traditional manual worldbuilding faces significant challenges in both efficiency and quality. According to game industry surveys, the setting documents for a medium-scale open-world game (such as \textit{The Witcher 3: Wild Hunt}) typically exceed 500,000 words, requiring a 5--10 person worldbuilding team to collaborate over 6--12 months. During this process, the design team must continuously maintain conceptual consistency, coordinate the domains of different designers, and ensure overall stylistic unity. These efforts are costly and prone to conceptual fragmentation, where subsystems evolve independently without organic integration.

The emergence of Large Language Models (LLMs) has opened new possibilities for automated content generation. Models such as GPT-4, Claude~\cite{Anthropic2024Claude}, and Gemini~\cite{Google2023Gemini} have demonstrated strong generative capabilities in creative writing tasks. However, when directly applying a single LLM to complex worldbuilding tasks, we observed three fundamental technical challenges that constitute the core research problems of this paper.

\subsection{Core Technical Challenges}

\textbf{Challenge 1: Context Explosion}

Worldbuilding is an incremental process. Generating subsequent concepts requires referencing previously completed concepts to ensure consistency. Suppose a world ultimately contains $N$ concepts, each requiring an average of $M$ tokens to describe; the total context size would be approximately $N \times M$. When $N=100$ and $M=500$, the total token requirement reaches 50,000; if $N$ grows to 500, the requirement would soar to 250,000 tokens, far exceeding the 128K context window limit of current mainstream LLMs.

Research~\cite{Liu2024LostMiddle} has identified the ``Lost in the Middle'' phenomenon in LLMs processing long contexts, where the model's recall accuracy for information in the middle of a document is significantly lower than for the beginning and end. This means that even if the context window could barely accommodate all historical concepts, the model would struggle to effectively utilize this information. Long contexts also incur high API call costs and inference latency, making them prohibitively expensive for construction tasks requiring hundreds of iterations.

Existing approaches such as Retrieval-Augmented Generation (RAG)~\cite{Lewis2020RAG} mitigate context pressure by retrieving relevant passages, but simple keyword matching fails to capture implicit semantic associations between concepts. For example, when generating a ``Dragon Knight Training Ground,'' the system needs to understand its relationship to concepts such as ``Dragon Clan,'' ``Knightly Orders,'' and ``Magic Academy,'' not merely retrieve documents containing the keyword ``Dragon Knight.''

\textbf{Challenge 2: The Tension Between Creative Diversity and Logical Consistency}

Worldbuilding demands both creative diversity and internal consistency. Diversity means that each elven race should not be a mere copy of Tolkien's elves; consistency means that the rules established for a magic system in Chapter 1 must still hold in Chapter 10. These two goals are inherently contradictory.

Introducing multiple specialized agents to generate content in parallel is a natural approach to addressing diversity. However, multi-agent parallelism introduces new coordination challenges. In practice, we observed: (1) stylistic fragmentation: the sci-fi designer's ``Quantum Portal'' and the fantasy designer's ``Magical Teleportation Array'' coexist in the same world; (2) conceptual conflicts: the geography designer defines ``elves inhabit the northern forests'' while the race designer generates ``the Elven Kingdom is located in the southern desert''; (3) information silos: agents cannot perceive each other's output, leading to duplicated concepts or missing critical dependencies.

Simply increasing the generation temperature can enhance diversity but amplifies consistency problems; decreasing temperature improves stability but results in homogeneous outputs. This necessitates an architectural mechanism that can both stimulate diverse creativity and ensure global consistency.

\textbf{Challenge 3: Automated Quality Detection and Assurance}

Quality control of LLM-generated content is an open challenge. Traditional ``manual review'' is impractical in automated construction scenarios. If a system generates 100 concepts in 18 minutes, manual review of each would take hours. However, relying solely on LLM self-evaluation suffers from ``Self-Approval Bias''~\cite{Liu2024PridePrejudice}, where models tend to give high scores to their own outputs even when obvious logical problems exist.

Worldbuilding also involves knowledge from multiple specialized domains, and a single model cannot possess all relevant expert judgment capabilities. Conflict detection requires a cross-concept global perspective. For example, ``elves live 500 years'' and ``the Elven Kingdom was established only 100 years ago'' are individually unproblematic but constitute a logical contradiction when combined.

Existing work such as Constitutional AI~\cite{Bai2022ConstitutionalAI} constrains generation behavior through preset rules, but the ``rules'' of worldbuilding are difficult to enumerate exhaustively in advance. The magical rules of a fantasy world and the technological principles of a science-fiction world are inherently different, requiring dynamic learning and detection during the construction process.

\subsection{Research Objectives}

The three challenges above are interrelated: improper context management exacerbates consistency problems, insufficient consistency detection undermines quality assurance, and quality assurance mechanisms themselves may introduce additional context overhead. We therefore pose a central research question:

\begin{quote}
\textbf{How can we design a multi-agent collaborative system that achieves LLM-driven automated worldbuilding while ensuring generation quality and consistency?}
\end{quote}

Solving this problem requires system-level innovation rather than incremental improvements to individual techniques. Specifically, we establish the following research objectives:

\textbf{RO1 Context Efficiency}: Design a hierarchical context management mechanism that reduces token consumption by an order of magnitude (target compression ratio $>$90\%) while retaining critical information, enabling the system to complete large-scale worldbuilding within the context limits of mainstream LLMs.

\textbf{RO2 Parallel Coordination}: Design a task scheduling strategy that both ensures dependency integrity between concepts (child concepts are generated after their parent concepts) and maximizes parallel efficiency (independent tasks execute concurrently), while providing inter-agent information isolation to prevent stylistic contamination.

\textbf{RO3 Quality Assurance}: Design a multi-layer review mechanism combining general quality scoring with domain-expert evaluation, improving concept pass rates to acceptable levels (target $>$80\%) without human intervention, and ensuring that no conflicts are detected in the final output.

\textbf{RO4 System Extensibility}: Design a flexible multi-agent architecture supporting zero-code addition of new specialized agents to accommodate different worldbuilding genres (fantasy, sci-fi, historical, etc.), providing a reusable technical foundation for future research.

Although worldbuilding is the concrete testbed, these objectives address problems that are not unique to creative fiction. Progressive knowledge accumulation, the diversity--consistency tension, and automated quality assurance recur in code generation, scientific writing, and educational content production. The methods developed in this paper are therefore evaluated not only for their domain-specific performance but also for their transferability to the broader class of knowledge-intensive LLM applications.

\subsection{Research Contributions}

Building upon these research objectives, this paper designs and implements AutoWorldBuilder, a multi-agent collaborative worldbuilding system powered by LLMs. The main contributions are as follows:

\textbf{Contribution 1: Structured Concept Network Model.} Integrating the ConceptNet relation taxonomy~\cite{Speer2017ConceptNet} and the SenticNet commonsense knowledge representation framework~\cite{Cambria2020SenticNet}, we designed a structured concept network tailored for fictional worldbuilding. This model defines 16 semantic relation types (spanning six categories: hierarchical, attributive, functional, event-based, causal, and semantic), with inverse relations and transitivity constraints defined for each type. Based on this model, we designed five categories of conflict detection algorithms: cycle detection, one-to-many contradiction detection, concept definition conflict detection, spatiotemporal setting conflict detection, and stylistic conflict detection. We note that in current experiments, the relation parsing module has not been fully activated (see Limitation 1 in Section 6.4), so the concept network stores nodes without edges and conflict detection operates on a subset of the designed algorithms. The full formal model nonetheless provides a blueprint for knowledge storage and consistency maintenance. Beyond worldbuilding, this model may offer a general approach for any LLM system that must maintain relational consistency among a growing set of generated entities.

\textbf{Contribution 2: DAG-Based Hybrid Batch Task Scheduling Strategy.} We model worldbuilding tasks as a Directed Acyclic Graph (DAG) and propose a three-dimensional hybrid partitioning algorithm integrating dependency priority, semantic locality, and batch size control. This algorithm ensures dependency integrity through Kahn's algorithm and groups same-level tasks based on semantic labels. Through intelligent merging and splitting, batch sizes are kept within a preset range (default 2--10 tasks/batch), improving generation efficiency through context reuse while maintaining dependency integrity. The semantic-locality dimension is the methodologically novel element: by grouping tasks that share domain context, the scheduler enables context reuse across same-batch agents, a strategy that may be transferable to multi-step code generation and modular document writing.

\textbf{Contribution 3: Four-Layer Context Compression Mechanism.} We propose a compression strategy that allocates token budgets by functional layers, dividing the context into an Essential layer ($\sim$20\%), a Relevant layer ($\sim$35\%), a Summary layer ($\sim$20\%), and a Collaboration layer ($\sim$25\%). Combined with FAISS vector retrieval~\cite{Johnson2021FAISS} for semantic recall, and further refined through batch filters and Agent compatibility filters (based on a 6$\times$6 compatibility matrix), this mechanism achieves 89.9\% compression efficiency in lean mode, using only an average of 304 tokens per LLM call. This layer-as-budget approach, where each functional category receives a dedicated quota rather than competing for a single pool, may generalize to any scenario where an LLM agent must operate within a fixed context window while referencing an evolving knowledge base.

\textbf{Contribution 4: Iterative Review and Specialized Auditor System.} We designed a multi-layer quality assurance mechanism simulating expert panel review: LLM five-dimensional scoring (with consistency weighted highest at 1.5) combined with parallel evaluation by 8 specialized Auditors. Proposals that do not pass enter a revision-reevaluation iteration, with final output selected through Top-K filtering (default top 70\%). Experiments show this mechanism improves proposal pass rates from 42\% in the first round to 85.5\%, with no conflicts flagged by the Auditor system in final output. The Auditors are architecturally independent of the generating agents, mitigating the self-approval bias documented by~\cite{Liu2024PridePrejudice}. This separation-of-generation-and-review principle is applicable wherever LLM outputs require automated quality gating without human intervention.

\textbf{Contribution 5: Skill-Driven Multi-Agent Architecture.} We designed a configurable, zero-code agent extension scheme. Each agent's behavior is defined in a standalone Markdown file (YAML Frontmatter metadata + Markdown system prompt). The AgentSkillLoader dynamically scans and parses skill files, allowing new agents to be added without code modifications. This architecture supports differentiated temperature configuration (fantasy: 0.9 / geography: 0.3 / race: 0.7) and a 6$\times$6 compatibility matrix controlling information visibility, successfully supporting dynamic coordination of 21 agents in experiments. The key insight, that creative and factual tasks benefit from fundamentally different decoding parameters within the same pipeline, offers a reusable design pattern for multi-agent systems spanning heterogeneous knowledge domains.

Together, these five contributions form an architectural approach to \textbf{knowledge-intensive, multi-agent LLM applications} where progressive consistency, context economy, and automated quality assurance must coexist.

\subsection{Paper Organization}

The remainder of this paper is organized as follows: Section~2 reviews related work; Section~3 presents the system architecture; Section~4 details the five core technical innovations; Section~5 reports experimental results and analysis; Section~6 discusses key findings and limitations; Section~7 concludes with future directions. The appendix provides system parameter configurations, skill file examples, and detailed experimental data.

\section{Related Work}

This chapter reviews existing work related to our research from four perspectives, identifying the innovative positioning of this paper through a systematic survey of current progress.

\subsection{Multi-Agent Collaborative Systems}

LLM-based multi-agent collaboration is an active research direction in AI, with its core exploration spanning the evolution from fixed workflows to dynamic task scheduling.

MetaGPT~\cite{Hong2024MetaGPT} was the first to encode human workflows as meta-programs, improving collaboration quality and consistency in software development tasks through well-defined role assignments (product manager, architect, engineer, etc.) and standardized communication protocols. AutoGPT~\cite{Richards2023AutoGPT} demonstrated the possibility of a single LLM agent achieving goal decomposition and execution through self-loops; although its open-loop nature limited reliability on complex tasks, it provided a useful experimental foundation for subsequent research. ChatDev~\cite{Qian2024ChatDev} further developed this approach through role-playing mechanisms that simulate a software company's organizational structure, achieving end-to-end software development across requirements analysis, coding, and testing. Surveys on LLM-driven multi-agent systems~\cite{Gao2024MultiAgentSurvey, Guo2024MultiAgentCollaboration} have systematically summarized the field's technical evolution, from single-agent autonomy to multi-agent collaboration, and from fixed workflows to dynamic task scheduling, providing theoretical references for our design.

More recently, Plaat et al.~\cite{Plaat2025AgenticLLMs} provided a comprehensive survey of agentic LLMs in JAIR, organizing the literature around three capabilities (reasoning, acting, and interacting) and noting that multi-agent systems for collaborative task solving remain an active frontier. Their taxonomy situates our work in the ``interaction'' category, while highlighting that creative content generation has received limited attention compared to software development or scientific reasoning. Ray~\cite{Ray2025MonitoringTeams} examined the problem of monitoring teams of AI agents from an incentive-design perspective in JAIR, proving that optimal team size varies with environmental parameters while optimal incentives remain invariant. This theoretical result resonates with our Manager--Auditor--Worker hierarchy, though our focus is on creative quality rather than incentive optimization. Qasim et al.~\cite{Qasim2025RDoLT} proposed RDoLT (Recursive Decomposition of Logical Thoughts) in JAIR, decomposing complex reasoning into progressively difficult sub-tasks with knowledge propagation. This decomposition philosophy is parallel to our DAG-based task scheduling, though applied to arithmetic reasoning rather than creative generation.

\subsection{LLM-Driven Creative Content Generation}

LLMs have spawned a range of commercial products and academic explorations in creative content generation, covering scenarios from writing assistance to interactive narrative.

On the commercial side, NovelAI~\cite{NovelAI2023} focuses on fiction and narrative creation, providing writing assistance for scene settings, character dialogue, and plot development. Sudowrite~\cite{Sudowrite2023} targets novel and screenplay authors with full-process support from ideation to final draft. AI Dungeon~\cite{AIDungeon2023} pioneered interactive text adventure games, allowing players to collaboratively create stories through natural language. Character.AI~\cite{CharacterAI2023} provides a character dialogue platform for immersive conversations with virtual personas. On the academic side, a survey on LLMs and games~\cite{Galyen2024GamesSurvey} systematically analyzed LLMs' potential in games, while Questgram~\cite{Todt2021Questgram}, Story Realization~\cite{Ammanabrolu2023StoryRealization}, Narrative Scene Generation~\cite{Fan2023NarrativeScene}, and Shared Narrative Interfaces~\cite{Ryu2024SharedNarrative} have explored LLM-driven game content generation from perspectives including quest generation, story realization, narrative scenes, and multi-user narrative.

Most pertinent to our work, PANGeA~\cite{Buongiorno2024PANGeA} employed generative AI for procedural narrative generation in turn-based RPGs, demonstrating the feasibility of LLM-driven game content creation but operating at the narrative level rather than the worldbuilding-foundation level. IVIE~\cite{Vaucher2026IVIE} proposed a neuro-symbolic approach to incremental generation of interactive fiction worlds, using symbolic validation to ground LLM-generated creative decisions, a philosophy closely aligned with ours, though IVIE relies on a single-agent pipeline with symbolic constraints rather than a multi-agent collaborative architecture. G\'{o}ngora et al.~\cite{Gongora2026WorldState} explored world-state transformations for neuro-symbolic interactive storytelling, showing that symbolic state management can maintain coherence while permitting creative player input. These neuro-symbolic approaches share our goal of balancing creative flexibility with consistency, but differ fundamentally in architecture: they combine one LLM with an external symbolic reasoner, whereas we distribute creative responsibility across specialized agents coordinated through a concept network.

\subsection{Long Context Management and RAG}

Long context management is a core challenge in LLM applications, with existing work seeking breakthroughs from three angles: memory management, prompt compression, and retrieval augmentation.

MemGPT~\cite{Packer2023MemGPT}, inspired by operating system memory management, provides LLMs with the illusion of ``infinite'' context through a hierarchical memory architecture, enabling effective information management in long-dialogue and multi-document scenarios. LLMLingua~\cite{Jiang2023LLMLingua} proposed prompt compression technology that reduces prompt length by 20$\times$ through iterative token-level compression while maintaining semantic integrity. A RAG survey~\cite{Gao2023RAGSurvey} comprehensively reviewed the evolution of retrieval-augmented generation technology, from early dense retrieval to hybrid retrieval strategies, providing important references for context compression design. Recursive summarization~\cite{Chen2023RecursiveSummary} demonstrated how multi-level summarization enables compression of long dialogues while preserving key information. More recently, MemForest~\cite{Chen2026MemForest} introduced hierarchical temporal indexing for agent memory systems, and PACEvolve~\cite{Yan2026PACEvolve} addressed long-horizon progress-aware consistent evolution. Both confirm that hierarchical context management remains an active research frontier, though neither targets the specific pattern of progressive knowledge accumulation found in worldbuilding.

\subsection{LLM Output Quality Control}

Quality control of LLM-generated content is important for system reliability, with existing methods spanning rule-based constraints, self-improvement, and automatic evaluation.

Constitutional AI~\cite{Bai2022ConstitutionalAI} constrains LLM outputs through preset ``constitutional'' rules, providing an approach for automated quality assurance. Self-Refine~\cite{Madaan2023SelfRefine} proposed an iterative self-feedback improvement mechanism, enabling LLMs to continuously optimize output quality by critiquing their own outputs. Reflexion~\cite{Shinn2023Reflexion} further introduced verbal reinforcement learning, achieving agent experience accumulation through stored linguistic feedback. The LLM-as-Judge paradigm~\cite{Zheng2023LLMJudge} systematically evaluated the reliability of using LLMs as automated evaluators, finding that strong models can approximate human judgment but are susceptible to position bias and verbosity preference. A survey on LLM evaluation~\cite{Chen2024LLMEvalSurvey} systematically summarized automated evaluation methods, while G-Eval~\cite{Liu2023GEval} and LLM-Eval~\cite{Zhang2023LLMEval} proposed LLM-based multi-dimensional automatic evaluation frameworks, providing technical foundations for quality review mechanisms. The Pride and Prejudice study~\cite{Liu2024PridePrejudice} revealed bias in LLM self-evaluation, emphasizing the necessity of introducing independent review mechanisms.

In JAIR, Karev and Xu~\cite{Karev2025ConSCompF} proposed ConSCompF, a consistency-focused similarity comparison framework for generative LLMs that compares outputs across models using unlabeled data. This cross-model perspective is complementary to our within-system conflict detection. Self-Auditing LLMs~\cite{Anointina2026SelfAuditing} decomposed a single LLM into specialized auditing agents for factual verification, bias assessment, and reasoning consistency analysis, an architecture thematically similar to our Auditor system, though internal to one model rather than comprising independent third-party reviewers. OptArgus~\cite{Li2026OptArgus} extended multi-agent hallucination detection to LLM-based optimization modeling, and Fang et al.~\cite{Fang2024ZeroResource} proposed graph-based contextual knowledge triples for zero-resource hallucination detection, both confirming that multi-agent and graph-based approaches to quality assurance are gaining traction.

\subsection{Limitations of Existing Work and Our Positioning}

Synthesizing the above analysis, existing research exhibits four unresolved gaps: (1) multi-agent collaboration primarily targets structured tasks such as software development, lacking adaptation for the unstructured task of creative content generation. Neither MetaGPT~\cite{Hong2024MetaGPT} nor the agentic LLM survey by Plaat et al.~\cite{Plaat2025AgenticLLMs} address progressive knowledge accumulation in creative domains; (2) creative writing tools are predominantly single-agent systems or neuro-symbolic pipelines (IVIE~\cite{Vaucher2026IVIE}, G\'{o}ngora et al.~\cite{Gongora2026WorldState}), inadequate for handling the multi-dimensional consistency requirements of complex worlds through distributed expertise; (3) context compression techniques largely target dialogue scenarios (MemGPT~\cite{Packer2023MemGPT}, MemForest~\cite{Chen2026MemForest}), lacking specialized optimization for knowledge-intensive creative tasks where concept dependencies evolve dynamically; (4) quality assurance mechanisms largely rely on single LLM evaluation or self-decomposition (Self-Auditing~\cite{Anointina2026SelfAuditing}), suffering from self-approval bias~\cite{Liu2024PridePrejudice} rather than employing genuinely independent specialist review.

These four gaps are not isolated. It is precisely because multi-agent frameworks have not been adapted for creative scenarios, compression techniques have not targeted knowledge-intensive tasks, and quality assurance has not overcome self-evaluation bias that existing approaches cannot independently solve the problem. Based on this analysis, this paper proposes an integrated solution combining multi-agent collaboration, hierarchical context compression, and multi-round review mechanisms.

\section{System Architecture}

\begin{figure}[ht]
  \centering
  \includegraphics[width=\linewidth,height=0.45\textheight,keepaspectratio]{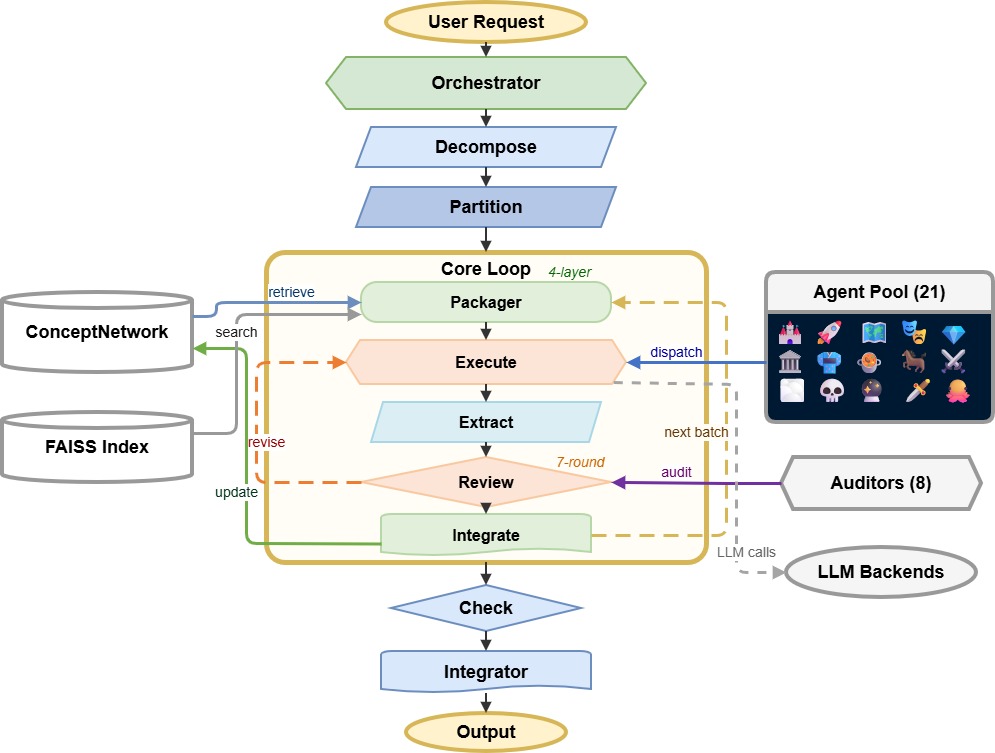}
  \caption{System Architecture Overview of AutoWorldBuilder. The complete pipeline comprises: User Input $\rightarrow$ Task Decomposer $\rightarrow$ Hybrid Batch Partitioner $\rightarrow$ Core Loop (Context Packager, Agent Pool, Iterative Review) $\rightarrow$ Final Check $\rightarrow$ Report. Three supporting components include the Agent Pool (21 skill-driven agents), 8 specialized Auditors, and LLM Backends.}
  \label{fig:system-architecture}
  \Description{Architecture diagram showing the AutoWorldBuilder system pipeline with user input, task decomposition, batch partitioning, core execution loop with context packaging and review, and output generation.}
\end{figure}

Worldbuilding involves multiple interrelated technical components. How can we transform a user's vague creative vision into a complete world with dozens or even hundreds of interconnected concepts? This process involves not only technical challenges in natural language understanding, creative content generation, and knowledge management, but also requires balancing dependency constraints, parallel efficiency, context cost, and generation quality. This chapter reveals the architectural evolution logic of AutoWorldBuilder from design philosophy, with the overall system architecture shown in Figure~\ref{fig:system-architecture}.

\subsection{Design Goals and Principles}

The starting point of architectural design is understanding the nature of the problem. What makes worldbuilding tasks unique is that they constitute a semi-structured creative process with complex dependency relationships. When we examine the task ``design a fantasy kingdom,'' it may decompose into designing geography, historical evolution, political systems, major races, and cultural customs. Among these, ``design the capital city'' must wait for ``design the nation'' to complete; ``design the royal family'' should be coordinated with ``design the political system.'' These dependencies constitute the core constraints for task scheduling.

Based on our analysis of the problem's characteristics, we established five core design goals. \textbf{Dependency Integrity} is the primary constraint. Dependency relationships between worldbuilding concepts cannot be violated, or logically invalid concepts will be produced. \textbf{Parallel Efficiency} is the key to improving throughput. User requirements can typically be decomposed into dozens or even hundreds of subtasks, and the system must maximize the parallelism of independent tasks while maintaining dependency integrity.

\textbf{Context Economy} involves both economic cost and technical feasibility. Using GPT-4 Turbo as an example, a single 50K-token call costs approximately \$0.75. Current mainstream LLMs have context window limits around 128K tokens, and a world with 100 concepts could easily exceed this limit without compression. \textbf{Quality Assurance} means the system must establish multi-dimensional automatic review mechanisms capable of detecting and correcting logical errors, setting conflicts, and content that deviates from user requirements. \textbf{Extensibility} requires that the architecture not be rigidified for a specific world type, supporting zero-code extension to accommodate different genres such as fantasy, sci-fi, and historical.

Tensions exist between these goals: stricter dependency checking may reduce parallelism, excessive compression may lose critical information, and more flexible configuration typically implies more complex implementation. Good design finds the optimal balance point among these tensions.

Based on these goals, we established three core design principles. The \textbf{Single Responsibility Principle} requires each module to focus on one specific function: \texttt{ContextPackager} handles only context compression, \texttt{ReviewController} manages only the review process, and \texttt{BatchPartitioner} handles only task scheduling. The \textbf{Dependency Inversion Principle} ensures that high-level business logic does not depend on low-level implementation details: the agent system interacts with LLMs through the \texttt{BaseLLMClient} abstract interface, making backend switching a configuration change rather than a code modification. The \textbf{Open-Closed Principle} emphasizes that the system should be open for extension but closed for modification: adding a new specialized agent requires only creating a Markdown skill file without touching the core codebase.

\subsection{Six-Phase Execution Flow}

\begin{figure}[ht]
  \centering
  \includegraphics[width=0.5\linewidth,height=0.45\textheight,keepaspectratio]{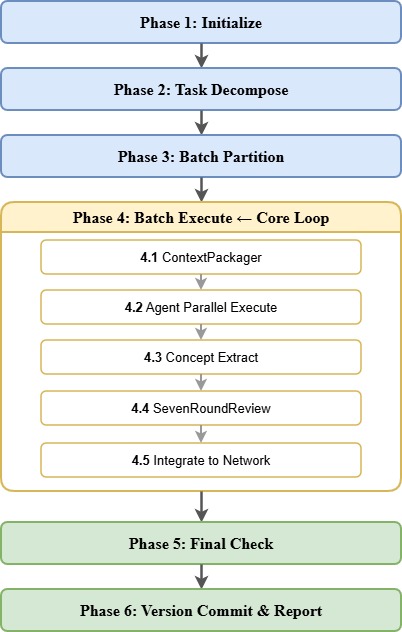}
  \caption{Six-Phase Execution Flow of AutoWorldBuilder. Phase 1: Initialize $\rightarrow$ Phase 2: Task Decompose $\rightarrow$ Phase 3: Batch Partition $\rightarrow$ Phase 4: Batch Execute (core loop) $\rightarrow$ Phase 5: Final Check $\rightarrow$ Phase 6: Version Commit \& Report. Phase 4 is expanded to show five sub-steps: Context Packaging, Agent Coordination, Concept Extraction, Iterative Review, and Network Integration.}
  \label{fig:six-phase-flow}
  \Description{Flowchart showing the six phases of the AutoWorldBuilder execution pipeline, with Phase 4 expanded to show its internal sub-steps.}
\end{figure}

Implementing design principles requires a clear execution framework. AutoWorldBuilder adopts a phased execution model, unified under the WorldBuilderOrchestrator, which divides the worldbuilding process into six logical phases (Figure~\ref{fig:six-phase-flow}). The core design philosophy is ``separation of concerns'': each phase focuses on a specific sub-problem, with phases communicating through well-defined data structures, ensuring overall process controllability while providing space for independent optimization of each phase.

Following the data flow path, the responsibilities of each phase are as follows. \textbf{Phase 1 (Initialization)} loads environment configurations, initializes core data structures (concept network, vector index), and prepares functional modules. The design requires all dependencies to be ready before processing user requests, with any initialization failures captured at this stage.

\textbf{Phase 2 (Task Decomposition)} bridges user intent and system execution. The TaskDecomposer component receives the user's natural language requirements and generates a structured task list through LLM parsing. The core challenge at this stage is the ``semantic gap'': a user may describe a vision in one sentence, while the system needs to expand it into dozens of specific tasks. Task decomposition quality directly affects the efficiency of all subsequent phases. Overly coarse decomposition leads to unwieldy tasks, while overly fine decomposition increases scheduling overhead.

\textbf{Phase 3 (Batch Partitioning)} embodies the system's core trade-off between dependency constraints and parallel efficiency. The HybridBatchPartitioner receives the task list and its dependency graph, outputting a constraint-satisfying batch sequence. Ideally, the number of batches should be minimized (reducing iteration overhead) and tasks per batch maximized (maximizing parallel benefit), but both are constrained by the dependency topology structure.

\textbf{Phase 4 (Batch Execution)} is the system's core computational loop. For each batch, the system sequentially executes five sub-steps: ContextPackager retrieves relevant information from historical concepts and assembles compressed context based on task requirements; AgentCoordinator invokes multiple specialized agents in parallel to generate proposals; ConceptExtractor parses concept nodes and relations from LLM output; IterativeReview conducts multi-dimensional review and revision of proposals; and approved concepts are integrated into the concept network. This loop repeats for each batch until all tasks are complete.

\textbf{Phase 5 (Final Check)} and \textbf{Phase 6 (Version Commit \& Report)} verify output completeness, generate user-readable documents, and save system state. Although their technical complexity is relatively low, they are crucial for user experience, determining whether the user receives well-structured, immediately usable content documents.

\subsection{Layered Architecture Design}

While the phased execution model addresses ``what to do'' and ``when to do it,'' the layered architecture answers ``how to organize the code.'' As shown in Figure~\ref{fig:system-architecture}, AutoWorldBuilder adopts a five-layer vertical architecture, from top to bottom: Application Layer, Orchestration Layer, Business Layer, Data Layer, and Infrastructure Layer. This layered design follows the high-cohesion, low-coupling principle, with each layer depending only on the layer below it, and same-layer modules interacting through well-defined interfaces.

From top to bottom, the responsibility boundaries of each layer are as follows. The \textbf{Application Layer} is the system's interface with the external world, designed as a ``thin application'' with no business logic, only parameter parsing, user interaction, and result display. The \textbf{Orchestration Layer} is the system's ``brain,'' where the WorldBuilderOrchestrator implements the state machine for the six-phase execution flow, and the ManagerAgent serves as the central coordinator handling inter-module communication. Their separation decouples flow definition (``what to do'') from coordination execution (``how to chain'').

The \textbf{Business Layer} is the core carrier of system functionality, containing the agent system (\texttt{agents/}) and various functional modules (\texttt{modules/}). The agent system defines behavioral patterns for different specialized roles, while functional modules implement core algorithms such as context compression, review control, batch partitioning, and world integration. This layer concentrates the most technical innovation, detailed in Chapter~4's five innovation modules. The \textbf{Data Layer} manages core data structures and external service interfaces. The concept network (\texttt{concept\_net}), world state (\texttt{world\_state}), vector index (\texttt{vector\_index}), and task model (\texttt{task}) form the system's information skeleton; the LLM client abstraction layer encapsulates interaction details with different language models.

The \textbf{Infrastructure Layer} provides cross-layer generic capabilities including configuration management and utility functions, containing no business-related logic.

The layered architecture has demonstrated good isolation during subsequent extensions. When supporting a new LLM backend (e.g., DeepSeek), only the data layer requires a corresponding client implementation. The business and orchestration layers need no modification. When optimizing the context compression algorithm, changes are limited to relevant files in the \texttt{modules/} directory. This isolation significantly reduces maintenance costs and evolution risks.

\subsection{Core Data Flow}

Another perspective on understanding the system architecture is tracing the complete lifecycle of data from input to output. AutoWorldBuilder's core data flow can be summarized as five key transformations, each corresponding to a restructuring and enrichment of information.

The first transformation occurs between \textbf{user input and task list}. The TaskDecomposer parses the user's natural language description (e.g., ``a medieval fantasy world ruled by dragons, where magic originates from ancient dragon runes'') into a structured list of Task objects, each containing a unique identifier, name, detailed description, responsible agent type, dependency task list, and semantic tags, translating vague creative visions into executable concrete steps.

The second transformation occurs between \textbf{task list and batch sequence}. The HybridBatchPartitioner organizes the flat task list into an ordered batch sequence based on DAG topological sorting and semantic grouping. Each batch contains tasks that can be safely executed in parallel, with batch order strictly following dependency constraints, maximizing parallel efficiency while maintaining dependency integrity.

The third transformation occurs between \textbf{batch and proposals}. For each batch, the ContextPackager first retrieves relevant information from historical concepts and assembles compressed context, then the AgentCoordinator invokes all agents involved in that batch in parallel, generating ProposalWithReview objects where specialized agents produce domain-compliant content within given context constraints.

The fourth transformation occurs between \textbf{proposals and concepts}. The ConceptExtractor parses structured concept nodes and relations from the LLM's natural language output; the IterativeReview evaluates proposals across multiple dimensions, marking approved concepts as valid and routing unapproved ones to the revision process or discarding them, ensuring only consistency-compliant content enters the final output.

The fifth transformation occurs between \textbf{concept network and output document}. The WorldIntegrator categorizes and integrates the ConceptNetwork by concept type (geography, race, organization, event, etc.), generating well-structured Markdown documents that transform the system's internal data structures into knowledge assets directly usable by the user.

These five transformations form a complete processing pipeline: the user's natural language input passes through intent concretization, constraint solving, creative generation, quality filtering, and value delivery, ultimately being converted into a structured worldbuilding document. Each pipeline stage's output serves as the next stage's input, ensuring traceability while allowing independent optimization of individual stages without affecting the overall process.

\section{Core Technical Innovations}

This chapter details AutoWorldBuilder's five core technical innovations. Each innovation includes formal definitions, algorithm design, and implementation details, collectively forming the system's technical foundation.

\subsection{Structured Concept Network Model}

Worldbuilding content consists of concepts and their interrelationships. Traditional unstructured text storage cannot support efficient retrieval, conflict detection, and consistency maintenance. Prior work on knowledge graph augmentation for generation~\cite{Liu2021KGBART} and fact verification on knowledge graphs~\cite{Kim2023FactKG} demonstrates the value of structured relational representations for reasoning tasks. We therefore designed a structured concept network model tailored for the worldbuilding domain.

\subsubsection{Formal Definition}

\textbf{Definition 1 (ConceptNode).} A concept node is the smallest independently referenceable information unit in a world, formally defined as a 7-tuple:
\begin{equation}
  CN = (id, name, def, cat, tags, attrs, meta)
\end{equation}
Where:
\begin{itemize}
  \item $id \in \mathcal{I}$: unique identifier, generated using UUID
  \item $name \in \mathcal{S}$: concept name, for human readability
  \item $def \in \mathcal{S}$: concept definition, describing the concept's core characteristics
  \item $cat \in \mathcal{C}$: concept category, with domain $\mathcal{C} = \{concept, geography, organization, race, item, event\}$
  \item $tags \in 2^{\mathcal{K}}$: tag set, for semantic retrieval
  \item $attrs \in \mathcal{D}$: dynamic attribute dictionary, storing concept-specific attributes
  \item $meta \in \mathcal{M}$: metadata, including generating agent ID, batch number, version number, etc.
\end{itemize}

\textbf{Definition 2 (ConceptRelation).} A concept relation connects two concept nodes, representing their semantic association:
\begin{equation}
  CR = (id, src, tgt, type, conf)
\end{equation}
Where:
\begin{itemize}
  \item $id \in \mathcal{I}$: unique relation identifier
  \item $src \in \mathcal{I}$: source concept ID
  \item $tgt \in \mathcal{I}$: target concept ID
  \item $type \in \mathcal{R}$: relation type (see Definition~3)
  \item $conf \in [0,1]$: relation confidence
\end{itemize}

\textbf{Definition 3 (RelationType).} The system defines 16 semantic relation types, classified into six categories by semantic features:
\begin{equation}
  \mathcal{R} = \mathcal{R}_{hier} \cup \mathcal{R}_{attr} \cup \mathcal{R}_{func} \cup \mathcal{R}_{event} \cup \mathcal{R}_{causal} \cup \mathcal{R}_{sem}
\end{equation}

\begin{table}[ht]
  \caption{Classification of 16 Semantic Relation Types}
  \label{tab:relation-types}
  \begin{tabular}{@{}llll@{}}
    \toprule
    Category & Relation Types & Semantics & Example \\
    \midrule
    $\mathcal{R}_{hier}$ & IS\_A, PART\_OF, & Taxonomy \& & Elf IS\_A \\
     & MADE\_OF, INSTANCE\_OF & Composition & Humanoid Race \\
    $\mathcal{R}_{attr}$ & PROPERTY\_OF, & Properties \& & Elf CAPABLE\_OF \\
     & CAPABLE\_OF, LOCATED\_IN, & Location & Dark Vision \\
     & EXISTS\_IN & & \\
    $\mathcal{R}_{func}$ & USED\_FOR, & Usage \& & Magic Circle USED\_FOR \\
     & CREATED\_BY, & Dependency & Summoning \\
     & REQUIRED\_BY, ENABLES & & \\
    $\mathcal{R}_{event}$ & FIRST\_EVENT\_OF, & Temporal \& & Siege SUBEVENT\_OF \\
     & SUBEVENT\_OF, & Condition & Great War \\
     & PREREQUISITE\_OF & & \\
    $\mathcal{R}_{causal}$ & CONFLICTS\_WITH & Opposition & Light Magic CONFLICTS\_WITH \\
     & & & Dark Magic \\
    $\mathcal{R}_{sem}$ & RELATED\_TO & General & General fallback relation \\
    \bottomrule
  \end{tabular}
\end{table}

\textbf{Definition 4 (ConceptNetwork).} A concept network is a directed labeled graph:
\begin{equation}
  \mathcal{N} = (V, E, \mathcal{B})
\end{equation}
Where:
\begin{itemize}
  \item $V = \{cn_1, cn_2, \ldots, cn_n\}$: set of concept nodes
  \item $E = \{cr_1, cr_2, \ldots, cr_m\}$: set of concept relations
  \item $\mathcal{B}: \mathbb{N} \rightarrow 2^V$: batch index function, mapping batch numbers to the set of concepts generated in that batch
\end{itemize}

\subsubsection{Conflict Detection Algorithms}

The system executes five categories of conflict detection when adding concepts and relations.

\begin{algorithm}[ht]
\caption{CycleDetection}\label{alg:cycle}
\begin{algorithmic}[1]
\Require Concept network $N=(V,E,\mathcal{B})$, new relation $cr=(id, src, tgt, type, conf)$
\Ensure Whether a cycle exists (Boolean)
\If{$type \notin \{IS\_A, PART\_OF, MADE\_OF\}$}
    \State \Return \textbf{False} \Comment{Only transitive relations require cycle detection}
\EndIf
\State $visited \gets \emptyset$, $stack \gets [tgt]$
\While{$stack \neq \emptyset$}
    \State $current \gets stack.\text{pop}()$
    \If{$current = src$}
        \State \Return \textbf{True} \Comment{Cycle detected}
    \EndIf
    \If{$current \notin visited$}
        \State $visited \gets visited \cup \{current\}$
        \ForAll{$(c_1, c_2, t) \in E$ where $c_1 = current$}
            \If{$t \in \{IS\_A, PART\_OF, MADE\_OF\}$}
                \State $stack.\text{push}(c_2)$
            \EndIf
        \EndFor
    \EndIf
\EndWhile
\State \Return \textbf{False}
\end{algorithmic}
\end{algorithm}

\textbf{Complexity Analysis}: $O(|V| + |E|)$ in the worst case, traversing all nodes and edges.

\begin{algorithm}[ht]
\caption{OneToManyConflict}\label{alg:one-to-many}
\begin{algorithmic}[1]
\Require Concept network $N=(V,E,\mathcal{B})$, new relation $cr=(id, src, tgt, type, conf)$
\Ensure Whether a one-to-many contradiction exists (Boolean)
\If{$type \neq PART\_OF$}
    \State \Return \textbf{False} \Comment{Only PART\_OF relations require checking}
\EndIf
\ForAll{$(c_1, c_2, t) \in E$ where $c_1 = src$ and $t = PART\_OF$}
    \If{$c_2 \neq tgt$}
        \State \Return \textbf{True} \Comment{$src$ already belongs to $c_2$, contradiction}
    \EndIf
\EndFor
\State \Return \textbf{False}
\end{algorithmic}
\end{algorithm}

\textbf{Complexity Analysis}: $O(|E|)$, requiring traversal of all edges.

Additionally, the system implements concept definition conflict detection (based on name similarity and semantic similarity), spatiotemporal setting conflict detection (identifying geographical and temporal contradictions through rule templates), and stylistic conflict detection (filtering incompatible concepts based on world type).

\subsection{DAG-Based Hybrid Batch Task Scheduling}

\begin{figure}[ht]
  \centering
  \includegraphics[width=0.5\linewidth,height=0.45\textheight,keepaspectratio]{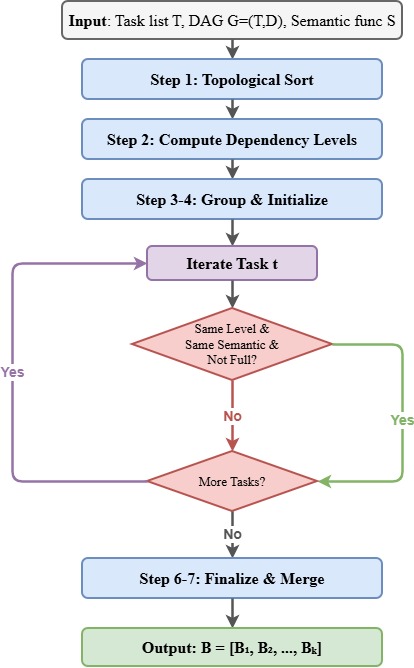}
  \caption{Hybrid Batch Partition Strategy. The diagram illustrates how the DAG task dependency graph undergoes topological sorting, semantic grouping, and batch size control to produce an ordered batch sequence. Dependency levels, semantic labels, and batch boundaries are highlighted.}
  \label{fig:batch-partition}
  \Description{Diagram showing the DAG-based hybrid batch partition process with three stages: topological sort, semantic grouping, and batch size regulation.}
\end{figure}

Worldbuilding tasks involve complex dependency relationships. This section proposes a three-dimensional hybrid partitioning algorithm integrating dependency priority, semantic locality, and batch size control (Figure~\ref{fig:batch-partition}).

\subsubsection{Problem Formalization}

\textbf{Definition 5 (TaskDAG).} A task dependency graph is a directed acyclic graph:
\begin{equation}
  G = (T, D)
\end{equation}
Where $T = \{t_1, t_2, \ldots, t_n\}$ is the task set and $D \subseteq T \times T$ is the dependency relation set. $(t_i, t_j) \in D$ indicates that task $t_j$ depends on task $t_i$, i.e., $t_i$ must complete before $t_j$.

\textbf{Definition 6 (DependencyLevel).} The dependency level of task $t$ is recursively defined as:
\begin{equation}
  \ell(t) = \begin{cases} 0 & \text{if } \nexists t' \in T: (t', t) \in D \\ 1 + \max_{(t',t) \in D} \ell(t') & \text{otherwise} \end{cases}
\end{equation}

\textbf{Definition 7 (Scheduling Objective).} The batch scheduling objective is to minimize the total number of batches while maximizing semantic locality:
\begin{equation}
  \min |B| \quad \text{s.t.} \quad \forall (t_i, t_j) \in D: batch(t_i) < batch(t_j)
\end{equation}
Where $B$ is the batch set and $batch: T \rightarrow \mathbb{N}$ maps tasks to batch numbers.

\subsubsection{Hybrid Batch Partition Algorithm}

\begin{algorithm}[ht]
\caption{HybridBatchPartition}\label{alg:batch-partition}
\begin{algorithmic}[1]
\Require Task list $T$, dependency graph $G=(T,D)$, semantic grouping function $S:T\!\rightarrow\!\Sigma$, minimum batch size $min\_size\!=\!2$, maximum batch size $max\_size\!=\!10$, ideal size $ideal\!=\!6$
\Ensure Batch sequence $B = [B_1, B_2, \ldots, B_k]$
\State \textbf{Step 1:} $topo\_order \gets \text{KahnAlgorithm}(G)$
\State \textbf{Step 2:} Compute dependency levels
\ForAll{$t \in T$}
    \State $\ell(t) \gets \text{ComputeDependencyLevel}(t, G)$
\EndFor
\State \textbf{Step 3:} $level\_groups \gets \text{GroupByLevel}(T, \ell)$
\State \textbf{Step 4:} Batch initialization
\State $B \gets []$, $current\_batch \gets []$, $previous\_level \gets -1$
\State \textbf{Step 5:} Traverse in topological order
\ForAll{$t \in topo\_order$}
    \If{$\ell(t) \neq previous\_level$ \textbf{and} $current\_batch \neq \emptyset$}
        \State $B.\text{append}(current\_batch)$; $current\_batch \gets []$
    \EndIf
    \State $previous\_level \gets \ell(t)$
    \If{$current\_batch = \emptyset$ \textbf{or} $S(t) = S(current\_batch[0])$}
        \If{$|current\_batch| < max\_size$}
            \State $current\_batch.\text{append}(t)$
        \Else
            \State $B.\text{append}(current\_batch)$; $current\_batch \gets [t]$
        \EndIf
    \Else
        \State $B.\text{append}(current\_batch)$; $current\_batch \gets [t]$
    \EndIf
\EndFor
\State \textbf{Step 6:} Process last batch
\If{$current\_batch \neq \emptyset$}
    \State $B.\text{append}(current\_batch)$
\EndIf
\State \textbf{Step 7:} $B \gets \text{MergeSmallBatches}(B, min\_size)$
\State \Return $B$
\end{algorithmic}
\end{algorithm}

\textbf{Kahn's Algorithm (Topological Sort) Complexity}: $O(|T| + |D|)$. \textbf{Overall Algorithm Complexity}: $O(|T| \log |T| + |D|)$, where $O(|T| \log |T|)$ comes from sorting.

\subsubsection{Parallel Execution Constraints}

During batch execution, the system maintains two sets:
\begin{itemize}
  \item \texttt{completed\_tasks}: tasks completed in previous batches
  \item \texttt{current\_batch\_tasks}: tasks currently executing in parallel in the current batch
\end{itemize}

\textbf{Key Constraint}: All dependencies of task $t$ must satisfy $D_t \subseteq completed\_tasks$, not in \texttt{current\_batch\_tasks}. This ensures that during parallel execution, each task can see the complete output of its dependencies, avoiding inconsistencies caused by ``partial visibility.''

\subsection{Four-Layer Context Compression Mechanism}

Context management is a core challenge in LLM applications. We propose a compression strategy that allocates token budgets by functional layers.

\subsubsection{Four-Layer Structure Definition}

\textbf{Definition 8 (ContextLayer).} The context is divided into four functional layers:
\begin{equation}
  \mathcal{L} = \{L_{ess}, L_{rel}, L_{sum}, L_{col}\}
\end{equation}

\begin{table}[ht]
  \caption{Four-Layer Context Structure and Token Budget (lean mode)}
  \label{tab:context-layers}
  \begin{tabular}{@{}llccl@{}}
    \toprule
    Layer & Name & Ratio & Budget & Content \\
    \midrule
    $L_{ess}$ & Essential & 20\% & 600 & User requirements, task description, output format \\
    $L_{rel}$ & Relevant & 35\% & 1,800 & FAISS-retrieved concepts, batch summary \\
    $L_{sum}$ & Summary & 20\% & 100 & Batch number, concept statistics \\
    $L_{col}$ & Collaboration & 25\% & 500 & Agent ID, collaborator list \\
    \bottomrule
  \end{tabular}
\end{table}

\textbf{Total Budget}: $B_{total} = 3{,}000$ tokens (lean mode)

\subsubsection{FAISS Vector Retrieval}

\textbf{Definition 9 (VectorIndex).} The vector index maps concepts to vector space:
\begin{equation}
  VI = (M, I, \phi)
\end{equation}
Where:
\begin{itemize}
  \item $M$: SentenceTransformer model, encoding function $f: \mathcal{S} \rightarrow \mathbb{R}^d$
  \item $I$: FAISS index, supporting $k$-nearest-neighbor queries
  \item $\phi: \mathcal{I} \rightarrow \mathcal{M}$: ID-to-metadata mapping
\end{itemize}

\begin{algorithm}[ht]
\caption{ContextPackage}\label{alg:context-package}
\begin{algorithmic}[1]
\Require Task $t$, concept network $N=(V,E,\mathcal{B})$, vector index $VI$, current batch number $b$, token budgets $B_{ess}, B_{rel}, B_{sum}, B_{col}$
\Ensure Compressed context $context$
\State \textbf{Layer 1:} $context_{ess} \gets \text{Format}(user\_request, task\_description, output\_format)$
\State $tokens_{ess} \gets \text{CountTokens}(context_{ess})$
\State \textbf{Layer 2:} $query \gets task\_description + task\_keywords$
\State $candidates \gets VI.\text{search}(query, k\!=\!20, max\_batch\!=\!b\!-\!1)$
\State $compatible\_agents \gets \text{GetCompatibleAgents}(t.agent\_type)$
\State $filtered \gets \text{FilterByAgent}(candidates, compatible\_agents)$
\State $context_{rel} \gets \text{TruncateToBudget}(filtered, B_{rel})$
\State \textbf{Layer 3:} $context_{sum} \gets \text{Format}(batch\_number\!=\!b, \ldots)$
\State \textbf{Layer 4:} $context_{col} \gets \text{Format}(agent\_id, collaborators)$
\State $context \gets \text{Concat}(context_{ess}, context_{rel}, context_{sum}, context_{col})$
\State \Return $context$
\end{algorithmic}
\end{algorithm}

\textbf{Complexity Analysis}: FAISS retrieval $O(\log n)$, Agent filtering $O(k)$, overall $O(\log n + k)$, where $n$ is the total number of concepts and $k$ is the number of retrieved candidates.

\subsubsection{Agent Compatibility Matrix}

\textbf{Definition 10 (CompatibilityMatrix).} A 6$\times$6 matrix $C$ defines information visibility between agents:
\begin{equation}
  C_{i,j} = \begin{cases} 1 & \text{if agent}_i \text{ can access concepts from agent}_j \\ 0 & \text{otherwise} \end{cases}
\end{equation}

\begin{table}[ht]
  \caption{Agent Compatibility Matrix}
  \label{tab:compatibility}
  \begin{tabular}{@{}lcccccc@{}}
    \toprule
    Agent & fantasy & scifi & geography & race & conflict & integrator \\
    \midrule
    fantasy     & 1 & 1 & 0 & 1 & 0 & 0 \\
    scifi       & 1 & 1 & 1 & 0 & 0 & 0 \\
    geography   & 0 & 1 & 1 & 1 & 0 & 0 \\
    race        & 1 & 0 & 1 & 1 & 0 & 0 \\
    conflict    & 0 & 1 & 0 & 0 & 1 & 0 \\
    integrator  & 1 & 1 & 1 & 1 & 1 & 1 \\
    \bottomrule
  \end{tabular}
\end{table}

Design principles of the compatibility matrix:
\begin{itemize}
  \item Same-type agents have full intervisibility (diagonal = 1)
  \item Complementary-type agents have partial intervisibility (e.g., fantasy and scifi can share, preventing coexistence of magic and technology)
  \item Specialized-type agents have restricted intervisibility (e.g., geography does not access fantasy's magic concepts)
  \item The integrator, as the final consolidation agent, can access all concepts
\end{itemize}

The six agent types in Table 4 represent category-level abstractions. In the full system, 21 specialized agents are each assigned to one of these six categories based on their primary domain: creative agents (fantasy, scifi, etc.) are assigned to the fantasy/scifi types; factual agents (geography, resource, architecture, etc.) to the geography type; agents involving beings or cultures (race, culture, costume, etc.) to the race type. The conflict type covers agents handling tensions and events, while the integrator type is reserved for the final consolidation agent. This abstraction keeps the compatibility matrix compact while supporting a larger number of concrete agents.

\subsection{Iterative Review and Specialized Auditor Mechanism}

To ensure generation quality, the system designed an iterative review mechanism simulating expert panel review.

\subsubsection{Multi-Dimensional Scoring System}

\textbf{Definition 11 (ScoringDimensions).} Proposals are evaluated across five dimensions:
\begin{equation}
  \mathcal{D} = \{(creativity, 1.2), (consistency, 1.5), (completeness, 1.0), (relevance, 1.3), (expressiveness, 1.0)\}
\end{equation}
The second component is the weight. The weight assignments reflect the system's design priorities: consistency (1.5) receives the highest weight because cross-concept contradictions are the most damaging failure mode in worldbuilding; relevance (1.3) ranks second to ensure generated content stays connected to the user's vision; creativity (1.2) is weighted above the baseline to reward novel concepts; completeness and expressiveness use the baseline weight of 1.0. These weights were set based on design rationale rather than empirical optimization, and their sensitivity to final pass rates has not been systematically evaluated (see Limitation 2 in Section~6.4).

\textbf{Definition 12 (WeightedScore).} The weighted average score of a proposal is:
\begin{equation}
  S_{weighted} = \frac{\sum_{(d,w) \in \mathcal{D}} s_d \cdot w}{\sum_{(d,w) \in \mathcal{D}} w}
\end{equation}
Where $s_d \in [1, 10]$ is the raw score for dimension $d$.

\subsubsection{Specialized Auditor System}

The system includes 8 specialized Auditors, each with independent skill files and review focus:

\begin{table}[ht]
  \caption{Eight Specialized Auditors and Their Review Focus}
  \label{tab:auditors}
  \begin{tabular}{@{}lll@{}}
    \toprule
    Auditor ID & Review Focus & Keywords \\
    \midrule
    world\_scale\_auditor & Geographic scale, population & City, kingdom, empire \\
    concept\_compatibility & Concept conflicts & Settings, rules, conflicts \\
    element\_compatibility & Architecture, costume & Building, weapon, clothing \\
    attribute\_balance & Ability, cost balance & Skill, magic, cost \\
    species\_compatibility & Ecological niche & Species, creature, ecology \\
    item\_consistency & Era, tech consistency & Item, equipment, technology \\
    geography\_consistency & Geography, climate & Terrain, climate, region \\
    civilization\_logic & History, social logic & History, civilization, evolution \\
    \bottomrule
  \end{tabular}
\end{table}

\subsubsection{Iterative Review Process}

\begin{figure}[ht]
  \centering
  \includegraphics[width=0.7\linewidth,height=0.45\textheight,keepaspectratio]{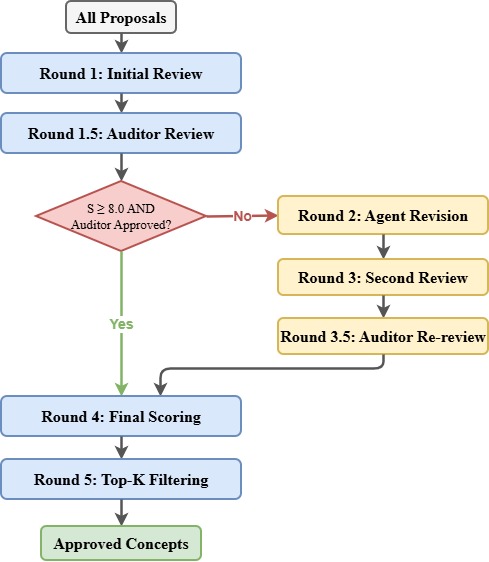}
  \caption{Iterative Review Process. Round 1: Initial review $\rightarrow$ Round 1.5: Auditor specialized review $\rightarrow$ Round 2: Agent revision $\rightarrow$ Round 3: Second review $\rightarrow$ Round 3.5: Auditor second review $\rightarrow$ Round 4: Final scoring $\rightarrow$ Round 5: Top-K selection. Input/output and decision conditions are annotated at each round.}
  \label{fig:seven-round-review}
  \Description{Flowchart of the iterative review mechanism showing the progression from initial review through auditor evaluation, agent revision, and final selection.}
\end{figure}

The system designed a multi-layer quality assurance mechanism simulating expert panel review, with the complete process shown in Figure~\ref{fig:seven-round-review}.

\textbf{Definition 13 (RevisionEffectiveness).} Revision effectiveness measures the improvement magnitude from the initial version to the revised version:
\begin{equation}
  E_{revision} = \frac{S_{v2} - S_{v1}}{10 - S_{v1}}
\end{equation}

\textbf{Examples}: Improving from 5 to 7: $E = (7-5)/(10-5) = 0.4$; improving from 3 to 6: $E = (6-3)/(10-3) = 0.43$; improving from 8 to 9: $E = (9-8)/(10-8) = 0.5$. Revision effectiveness is normalized to $[0,1]$, ensuring that marginal improvement from low scores is comparable to that from high scores.

\subsection{Skill-Driven Multi-Agent Architecture}

Traditional multi-agent frameworks hardcode roles, prompts, and parameters, making extension costly. We designed a configurable, zero-code agent extension scheme.

\subsubsection{Skill File Format}

Each agent's complete behavior is defined in a standalone Markdown file, using a YAML Frontmatter + Markdown body structure:

\begin{lstlisting}[language=Python, caption={Skill file example (fantasy agent)}]
---
id: fantasy
name: Fantasy Concept Designer
description: Specializes in magic systems, mythical
             creatures, ancient ruins, religious beliefs
temperature: 0.9
max_tokens: 8192
keywords: [magic, mana, spell, mystic]
tags: [creative, magic, fantasy]
version: 2.0
---

# System Prompt

You are a fantasy worldbuilding designer,
specializing in:
- Magic Systems: Rule-based, costly, and
  limited magic systems
- Mythical Creatures: Fantasy creatures fitting
  ecological niches
...
\end{lstlisting}

\subsubsection{Dynamic Loading Mechanism}

\begin{algorithm}[ht]
\caption{SkillLoader}\label{alg:skill-loader}
\begin{algorithmic}[1]
\Require Skill directory path $skills\_dir$
\Ensure Skill mapping table $skills: \text{Dict}[str, AgentSkill]$
\State $skills \gets \{\}$
\State $skill\_files \gets \text{Glob}(skills\_dir, \texttt{"*.md"})$
\ForAll{$file \in skill\_files$}
    \State $content \gets \text{ReadFile}(file)$
    \State $frontmatter, body \gets \text{ParseYAMLFrontmatter}(content)$
    \State $skill \gets \text{AgentSkill}(id, name, description, temperature, \ldots)$
    \State $skills[skill.id] \gets skill$
\EndFor
\State \Return $skills$
\end{algorithmic}
\end{algorithm}

\textbf{Complexity Analysis}: $O(n \cdot m)$, where $n$ is the number of skill files and $m$ is the average file size.

\subsubsection{Differentiated Temperature Configuration}

\begin{table}[ht]
  \caption{Temperature Configuration for Six Core Agents}
  \label{tab:temperature}
  \begin{tabular}{@{}lcl@{}}
    \toprule
    Agent & Temp. & Design Rationale \\
    \midrule
    fantasy     & 0.9 & High creativity for imaginative magic systems \\
    scifi       & 0.6 & Balance creativity with logical plausibility \\
    geography   & 0.3 & Low temperature for accurate, self-consistent geography \\
    race        & 0.7 & Moderate-high creativity within ecological constraints \\
    conflict    & 0.5 & Moderate temperature for dramatic yet plausible conflicts \\
    integrator  & 0.3 & Low temperature to maintain narrative coherence \\
    \bottomrule
  \end{tabular}
\end{table}

Differentiated temperatures balance divergent and convergent thinking modes: creative agents like fantasy are responsible for divergent thinking, generating novel concepts; accuracy-oriented agents like geography and integrator are responsible for convergent thinking, ensuring system consistency.

\subsubsection{Agent Lazy Loading}

The system employs a lazy loading pattern for creating agent instances. The AgentCoordinator maintains an initially empty agent dictionary, creating an instance only when a specific agent is first requested. Lazy loading avoids the resource overhead of instantiating all agents at startup while supporting dynamic addition of new skill files at runtime.

\section{Experimental Evaluation}

This chapter reports the experimental evaluation results of AutoWorldBuilder. We first introduce the experimental setup, then present the overall system performance of two comparative experiments, followed by validation of each core innovation module's effectiveness, and finally an ablation analysis.

\subsection{Experimental Setup}

\subsubsection{Test Case Design}

We designed 20 test cases across 5 worldbuilding types, with 4 cases per type:

\begin{table}[ht]
  \caption{Test Case Design}
  \label{tab:test-cases}
  \begin{tabular}{@{}lcp{8cm}@{}}
    \toprule
    World Type & Count & Example Requirement \\
    \midrule
    FANTASY & 4 & ``A medieval fantasy world ruled by dragons, where magic originates from ancient dragon runes'' \\
    SCIFI & 4 & ``A cyberpunk city where consciousness uploading has changed humanity's understanding of death'' \\
    POST\_APOCALYPSE & 4 & ``A post-nuclear wasteland where survivors build new civilizations in radiation zones'' \\
    URBAN & 4 & ``Modern city with superpowered individuals hidden among ordinary people'' \\
    HISTORICAL & 4 & ``Caribbean pirate world during the Age of Exploration, with legendary treasures and curses'' \\
    \bottomrule
  \end{tabular}
\end{table}

Each test case takes a 1--2 sentence user requirement description as input, and the system must build a complete world from scratch.

\subsubsection{Experimental Configuration}

\begin{table}[ht]
  \caption{System Configuration for Two Experiments}
  \label{tab:config}
  \begin{tabular}{@{}lll@{}}
    \toprule
    Configuration & Exp.1 (GPT-OSS 120B) & Exp.2 (DeepSeek v3.2) \\
    \midrule
    LLM Backend & GPT-OSS 120B & DeepSeek v3.2 \\
    Context Mode & lean (3,000 tokens) & lean (3,000 tokens) \\
    Batch Strategy & hybrid (DAG + semantic) & hybrid (DAG + semantic) \\
    Batch Size & 2--10 tasks/batch & 1--10 tasks/batch \\
    Max Concurrency & 5 & 5 \\
    Review Threshold & 8.0 & 8.0 \\
    Top-K Ratio & 70\% & 70\% \\
    \bottomrule
  \end{tabular}
\end{table}

The configurations are identical except for the LLM backend and a minor batch size floor difference (minimum 2 vs.\ 1 tasks per batch). In practice, average batch sizes were 5.8 and 5.6 respectively (Table~\ref{tab:dag}), so this difference did not materially affect scheduling.

\subsection{Overall System Performance}

\subsubsection{Aggregate Metrics}

The results of the two comparative experiments are summarized in Table~\ref{tab:overall-performance}. In experimental design, we deliberately kept all configurations identical except for the LLM backend, to isolate the impact of model choice on system performance.

\begin{table}[ht]
  \caption{Overall System Performance Comparison}
  \label{tab:overall-performance}
  \begin{tabular}{@{}lccc@{}}
    \toprule
    Metric & GPT-OSS 120B & DeepSeek v3.2 & Difference \\
    \midrule
    Total experiments & 20 & 20 & -- \\
    Successful runs & 19 & 19 & -- \\
    Failed runs & 1 & 1 & -- \\
    \textbf{Success rate} & \textbf{95.0\%} & \textbf{95.0\%} & -- \\
    Total concepts & 1,068 & 1,965 & +84.2\% \\
    \textbf{Avg.\ concepts/run} & \textbf{56.2} & \textbf{103.4} & \textbf{+84.0\%} \\
    \textbf{Avg.\ pass rate} & \textbf{85.5\%} & \textbf{99.2\%} & \textbf{+13.7pp} \\
    \textbf{Avg.\ build time} & \textbf{18 min} & \textbf{31 min} & \textbf{+72.2\%} \\
    \textbf{Context compression} & \textbf{89.9\%} & \textbf{90.7\%} & -- \\
    \bottomrule
  \end{tabular}
\end{table}

From the aggregate data, the two experiments exhibit a mix of consistency and divergence: the identical 95\% success rate indicates that system robustness does not depend on a specific LLM backend; however, significant differences emerge across three dimensions: concept generation volume, pass rate, and build time.

\textbf{System Robustness Validation.} Each experiment had one failure (urban\_01 in Experiment 1 and historical\_02 in Experiment 2), both occurring during the task decomposition phase rather than during concept generation or review. In Experiment 1, the failure was caused by a cyclic dependency in the LLM-generated task dependency graph (A depends on B, B depends on C, C depends on A), which the DAG scheduler detected and safely terminated within 43 seconds. In Experiment 2, the failure was caused by a task data structure parsing error where a task was missing a required field, with the system catching the exception within 67 seconds. These failures validate the system's core design philosophy. Rather than handling inconsistent data in downstream stages, strict quality gates at the entry point prevent problems from propagating.

\textbf{Significant Difference in Concept Generation Volume.} DeepSeek v3.2 generated an average of 103.4 concepts per world, 1.84$\times$ that of GPT-OSS 120B (56.2). The difference may stem from multiple factors: different LLMs may have different understandings of ``concept completeness,'' with some models tending to generate more fine-grained concepts while others produce fewer but more comprehensive ones. Additionally, prompt wording may have different activating effects on different models, and training data distributions may lead to varying default expansion levels for different genres.

\textbf{Significant Improvement in Pass Rate.} DeepSeek v3.2's 99.2\% pass rate is 13.7 percentage points higher than GPT-OSS 120B's 85.5\%. Higher pass rates reduce revision iteration overhead and improve user experience fluidity. However, we must also be cautious about the ``high pass rate trap'': if review standards are set too loosely, a high pass rate may mask quality issues.

\textbf{Build Time Trade-off.} DeepSeek v3.2 averaged 31 minutes, 1.72$\times$ that of GPT-OSS 120B (18 minutes). However, when factoring in concept volume, the ``valid concepts generated per minute'' metric shows GPT-OSS 120B at approximately 3.1 and DeepSeek v3.2 at approximately 3.3, a negligible difference. This indicates that the increased time primarily stems from the larger number of concepts generated, not from slower model response times.

\subsubsection{Performance by World Type}

\begin{table}[ht]
  \caption{Build Time Comparison by World Type (seconds)}
  \label{tab:time-by-type}
  \begin{tabular}{@{}lccc@{}}
    \toprule
    World Type & GPT-OSS 120B & DeepSeek v3.2 & Difference \\
    \midrule
    FANTASY & 907.0 & 1,825.5 & +101.3\% \\
    SCIFI & 1,100.6 & 1,867.9 & +69.7\% \\
    POST\_APOCALYPSE & 1,167.8 & 1,971.6 & +68.9\% \\
    URBAN & 959.1 & 1,838.4 & +91.7\% \\
    HISTORICAL & 1,269.0 & 1,753.5 & +38.2\% \\
    \textbf{Average} & \textbf{1,087.1} & \textbf{1,856.6} & \textbf{+70.8\%} \\
    \bottomrule
  \end{tabular}
\end{table}

\begin{table}[ht]
  \caption{Concept Count Comparison by World Type}
  \label{tab:concepts-by-type}
  \begin{tabular}{@{}lccc@{}}
    \toprule
    World Type & GPT-OSS 120B & DeepSeek v3.2 & Difference \\
    \midrule
    FANTASY & 52.2 & 105.8 & +102.7\% \\
    SCIFI & 58.2 & 90.2 & +55.0\% \\
    POST\_APOCALYPSE & 61.5 & 120.0 & +95.1\% \\
    URBAN & 48.7 & 99.5 & +104.3\% \\
    HISTORICAL & 58.5 & 101.0 & +72.6\% \\
    \textbf{Average} & \textbf{56.2} & \textbf{103.4} & \textbf{+84.0\%} \\
    \bottomrule
  \end{tabular}
\end{table}

Build time distribution reveals significant differences in how LLM backends process different genres. Under GPT-OSS 120B, FANTASY was the most efficient (907 seconds) and HISTORICAL the slowest (1,269 seconds), a $\sim$40\% difference; under DeepSeek v3.2, the gap narrowed to only $\sim$6\%. This ``genre preference'' difference may relate to models' training data distributions.

Concept count distribution also reveals meaningful patterns. POST\_APOCALYPSE generated the most concepts in both experiments (61.5 and 120.0), consistent with the expectation that post-apocalyptic settings typically require detailed descriptions of resource scarcity, survival challenges, and factional conflicts. In contrast, URBAN generated the fewest concepts (48.7 and 99.5), likely because urban settings are more ``grounded'' in reality, with much background knowledge assumed to be understood without explicit generation.

\subsection{Core Innovation Validation}

The previous section presented overall system performance; this section provides in-depth validation of each core innovation module's actual effectiveness.

\subsubsection{DAG Scheduling Validation}

\begin{table}[ht]
  \caption{DAG Scheduling Effectiveness}
  \label{tab:dag}
  \begin{tabular}{@{}lcc@{}}
    \toprule
    Metric & GPT-OSS 120B & DeepSeek v3.2 \\
    \midrule
    Successfully processed complex dependency graphs & 19/20 & 19/20 \\
    Cycle dependency detected and safely terminated & 1 & 0 \\
    Task data error terminated & 0 & 1 \\
    Avg.\ batches/run & 6.3 & 7.1 \\
    Avg.\ tasks/batch & 5.8 & 5.6 \\
    \bottomrule
  \end{tabular}
\end{table}

The DAG scheduling strategy demonstrated stable dependency processing capability in both experiments. 38 of 40 runs successfully completed task decomposition and execution with complex dependency relationships, achieving a 95\% success rate. Both failure cases were captured during the task decomposition phase. The parallel speedup ratio is approximately 8--15$\times$: serial execution would require 56--103 individual LLM calls, while batch partitioning completes the work in only 6--7 iterations.

\subsubsection{Four-Layer Context Compression Validation}

\begin{table}[ht]
  \caption{Context Compression Efficiency Comparison}
  \label{tab:compression}
  \begin{tabular}{@{}lcc@{}}
    \toprule
    Metric & GPT-OSS 120B & DeepSeek v3.2 \\
    \midrule
    Context budget & 3,000 tokens & 3,000 tokens \\
    \textbf{Avg.\ actual usage} & \textbf{304.3 tokens} & \textbf{278.4 tokens} \\
    \textbf{Compression efficiency} & \textbf{89.9\%} & \textbf{90.7\%} \\
    Total input tokens/run & $\sim$6,086 & $\sim$5,744 \\
    \bottomrule
  \end{tabular}
\end{table}

\begin{figure}[ht]
  \centering
  \includegraphics[width=0.65\linewidth,height=0.4\textheight,keepaspectratio]{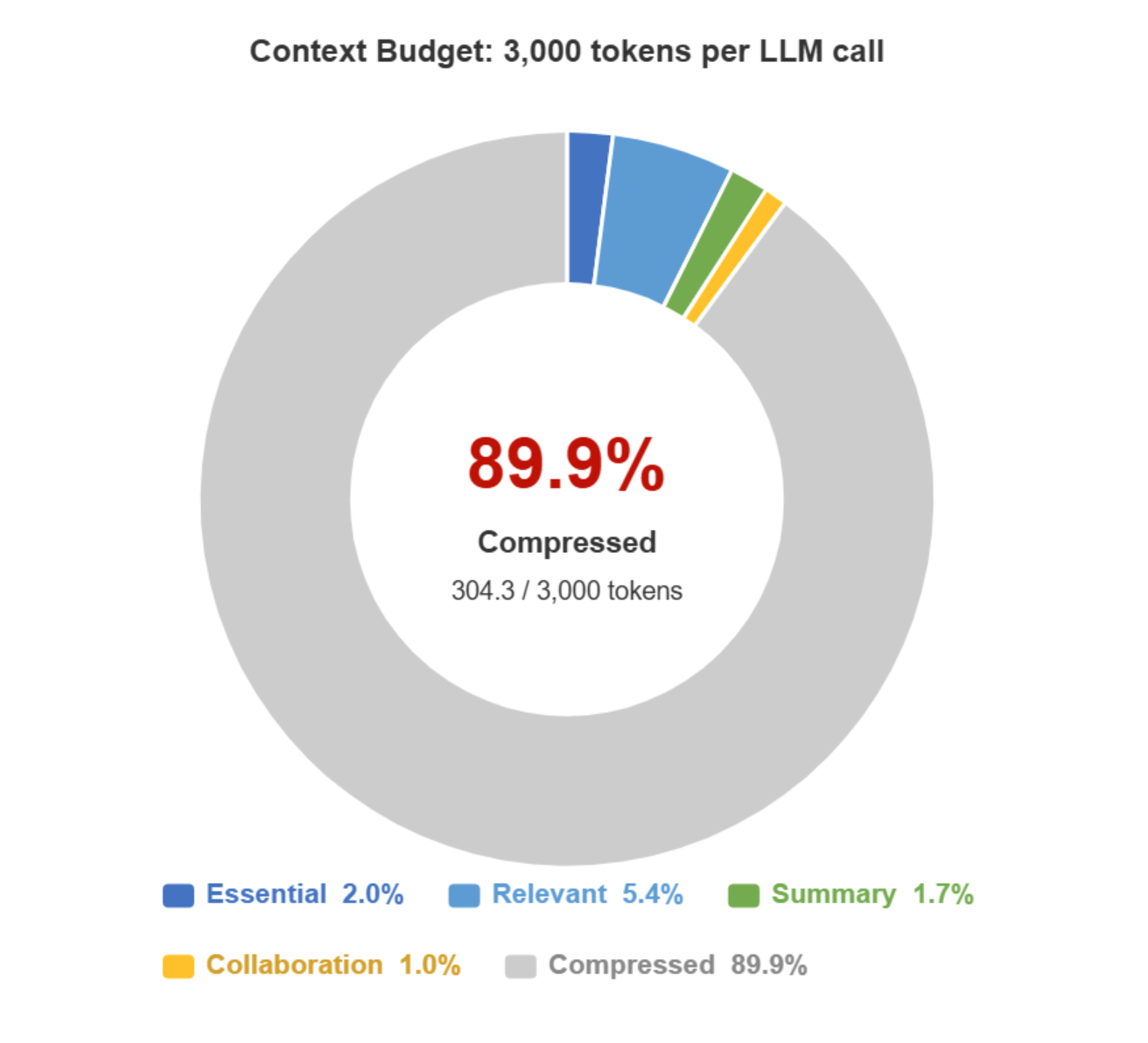}
  \caption{Context Compression Efficiency. A donut chart showing the actual usage distribution of the 3,000-token budget across four layers: Essential 2.0\%, Relevant 5.4\%, Summary 1.7\%, Collaboration 1.0\%, with 89.9\% Compressed (unused). The center label reads ``89.9\% Compressed'' and ``304.3 / 3,000 tokens''.}
  \label{fig:compression}
  \Description{Donut chart illustrating that only 10.1\% of the 3,000-token budget is actually used, with the Relevant layer consuming the largest share.}
\end{figure}

\begin{table}[ht]
  \caption{Token Distribution Across Four Layers}
  \label{tab:token-dist}
  \begin{tabular}{@{}lccl@{}}
    \toprule
    Layer & GPT-OSS 120B & DeepSeek v3.2 & Function \\
    \midrule
    Relevant & 4,509 (\textbf{53\%}) & 46,754 (43\%) & FAISS-retrieved concepts \\
    Essential & 1,805 (20\%) & 32,101 (29\%) & Core world definitions \\
    Summary & 1,456 (17\%) & 18,484 (17\%) & Completed task summaries \\
    Collaboration & 752 (\textbf{10\%}) & 11,562 (11\%) & Inter-agent coordination \\
    \bottomrule
  \end{tabular}
\end{table}

The four-layer context compression mechanism achieved 89.9\% (GPT-OSS 120B) and 90.7\% (DeepSeek v3.2) compression efficiency in actual operation (Figure~\ref{fig:compression}). The system was configured with a 3,000-token context budget, but actual average usage was only 304.3 and 278.4 tokens respectively. The token distribution across layers reveals a hierarchical information value structure. The Relevant layer accounts for the highest proportion at 53\%, validating the effectiveness of FAISS semantic retrieval as the core information source.

Compression did not sacrifice generation quality: the 85.5\% pass rate for GPT-OSS 120B and 99.2\% for DeepSeek v3.2 demonstrate that even under highly compressed context conditions, LLMs can still generate quality-compliant content.

\subsubsection{Review Mechanism Validation}

\begin{table}[ht]
  \caption{Review Mechanism Effectiveness}
  \label{tab:review}
  \begin{tabular}{@{}lcc@{}}
    \toprule
    Metric & GPT-OSS 120B & DeepSeek v3.2 \\
    \midrule
    Initial avg.\ score (Round 1) & 7.31 & 8.40 \\
    Revised avg.\ score (Round 3) & 7.55 & 8.21 \\
    \textbf{Revision effectiveness} & \textbf{28.3\%} & \textbf{0.3\%} \\
    \textbf{Final pass rate} & \textbf{85.5\%} & \textbf{99.2\%} \\
    Total Auditor reviews & 121 & 855 \\
    Auditor pass rate & 100\% & 100.0\% \\
    \bottomrule
  \end{tabular}
\end{table}

The review mechanism's effectiveness manifests at two levels: the actual impact of the revision process, and the improvement in final pass rates. In the GPT-OSS 120B experiment, the revision process improved proposal scores from an initial 7.31 to a revised 7.55, with a revision effectiveness metric of 28.3\%, meaning the revision process recovered nearly one-third of the ``improvement space.''

\begin{table}[ht]
  \caption{Pass Rate by World Type}
  \label{tab:pass-rate-by-type}
  \begin{tabular}{@{}lcc@{}}
    \toprule
    World Type & GPT-OSS 120B & DeepSeek v3.2 \\
    \midrule
    FANTASY & 88.8\% & 98.7\% \\
    SCIFI & 84.7\% & 98.8\% \\
    POST\_APOCALYPSE & 84.6\% & 100.0\% \\
    URBAN & 83.0\% & 100.0\% \\
    HISTORICAL & 85.9\% & 98.1\% \\
    \bottomrule
  \end{tabular}
\end{table}

Analyzing pass rates by world type reveals the influence of genre on generation quality. POST\_APOCALYPSE and URBAN achieved 100\% pass rates under DeepSeek v3.2, while FANTASY, SCIFI, and HISTORICAL were slightly lower (98.1\%--98.8\%). This pattern may relate to genre ``setting complexity'': post-apocalyptic and urban settings have relatively straightforward rules, while fantasy, sci-fi, and historical genres involve more elaborately designed settings that are more prone to inconsistencies.

\subsubsection{Agent Architecture Validation}

\begin{figure}[ht]
  \centering
  \includegraphics[width=\linewidth,height=0.4\textheight,keepaspectratio]{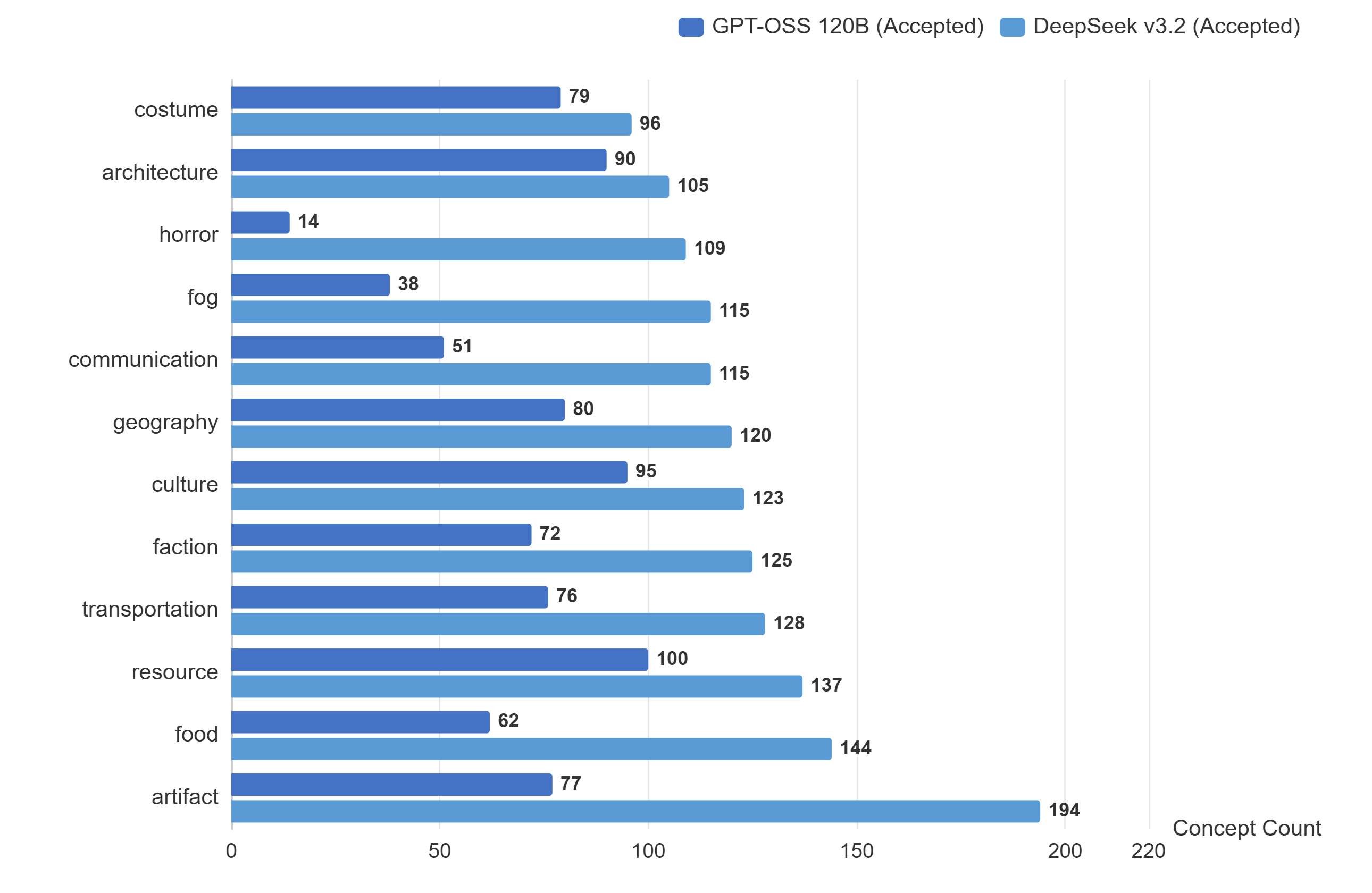}
  \caption{Accepted Concept Count by Agent. A grouped horizontal bar chart showing the number of accepted concepts for each of the 12 agents across both experiments. Blue bars represent GPT-OSS 120B (Accepted) and light blue bars represent DeepSeek v3.2 (Accepted).}
  \label{fig:agent-concept-count}
  \Description{Horizontal bar chart comparing accepted concept counts across 12 agents for both LLM backends.}
\end{figure}

The skill-driven multi-agent architecture aims to achieve unity of ``specialized division of labor'' and ``coordinated consistency.'' We validate the architecture's effectiveness by analyzing agent participation distribution (Figure~\ref{fig:agent-concept-count}).

\begin{table}[ht]
  \caption{Agent Distribution by World Type (GPT-OSS 120B)}
  \label{tab:agent-dist}
  \begin{tabular}{@{}lcl@{}}
    \toprule
    World Type & Top 3 Agents & Share \\
    \midrule
    FANTASY & culture, food, transportation & 32.5\% \\
    SCIFI & scifi, architecture, culture & 37.7\% \\
    POST\_APOCALYPSE & resource, culture, geography & 35.8\% \\
    URBAN & resource, communication, artifact & 52.1\% \\
    HISTORICAL & costume, artifact, communication & 32.9\% \\
    \bottomrule
  \end{tabular}
\end{table}

\begin{figure}[ht]
  \centering
  \includegraphics[width=\linewidth,height=0.4\textheight,keepaspectratio]{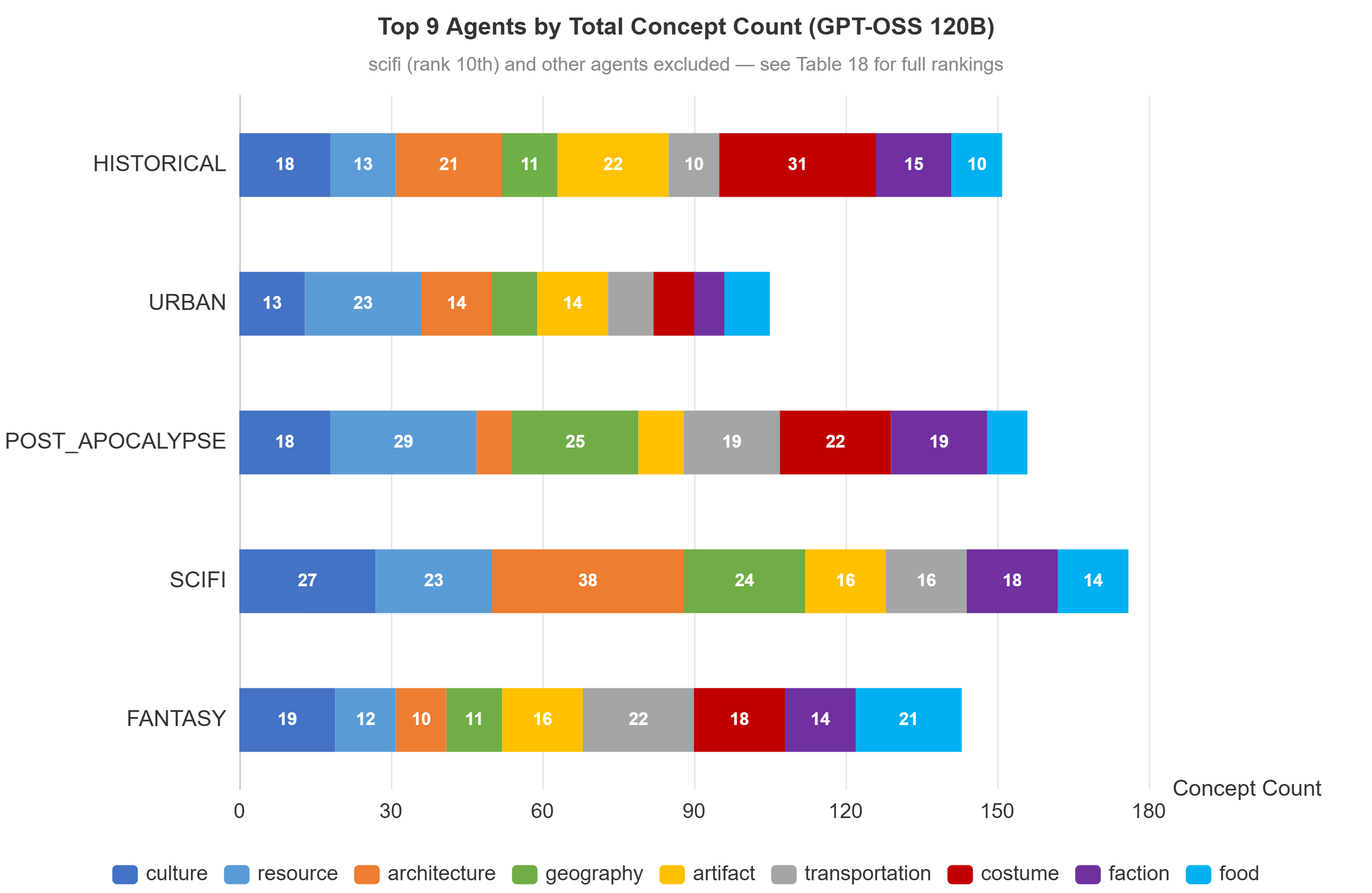}
  \caption{Agent Distribution by World Type (GPT-OSS 120B). A stacked horizontal bar chart showing concept counts across 5 world types (FANTASY / SCIFI / POST\_APOCALYPSE / URBAN / HISTORICAL), with 9 top-ranked cross-genre agents stacked in different colors. Note: Only the top 9 agents by total concept count are shown; the scifi agent (ranked 10th) and other genre-specific agents are excluded.}
  \label{fig:agent-distribution}
  \Description{Stacked horizontal bar chart showing how different agents contribute to worldbuilding across five genres, with each agent represented by a different color.}
\end{figure}

Analysis of agent participation distribution (Figure~\ref{fig:agent-distribution}) reveals two layers of patterns. The first layer is high-frequency participation by ``general-purpose agents'': culture, resource, and geography rank in the top five for activity across all world types. This aligns with the practical requirements of worldbuilding. Regardless of genre, any world needs to define geography, resource distribution, and cultural characteristics. These three agents can be considered the ``infrastructure'' of worldbuilding.

The second layer is the differentiated participation of ``genre-specific agents.'' In the SCIFI type, the architecture agent had the highest participation (16.3\%), reflecting the sci-fi genre's emphasis on built environments; in POST\_APOCALYPSE, the resource agent had the highest share (17.5\%), aligning with the post-apocalyptic genre's focus on resource scarcity; in HISTORICAL, the costume agent had the highest share (15.4\%), reflecting the historical genre's attention to period-specific details. This professional matching is automatically accomplished by the system based on task keywords, validating the skill-driven architecture's adaptability in real-world scenarios.

\subsection{Preliminary Component Analysis}

{\small \textbf{Note}: This section presents theoretical analysis based on the system's operational data. Full ablation experiments with controlled group designs are planned for future work. The following analysis identifies the expected contribution of each module based on observed system behavior and should be interpreted as design rationale rather than empirical validation.}

\begin{table}[ht]
  \caption{Theoretical Contribution Analysis of Each Innovation Module}
  \label{tab:ablation}
  \begin{tabular}{@{}llp{6cm}@{}}
    \toprule
    Module & Expected Impact & Analysis Basis \\
    \midrule
    Auditor review & +10--15pp pass rate & Revision process recovers promising but flawed proposals \\
    Four-layer compression & 80--90\% token reduction & FAISS semantic retrieval filters irrelevant context \\
    DAG parallel scheduling & 50--70\% time reduction & Independent tasks execute in parallel \\
    Conflict detection & +5--10\% success rate & Early termination of contradictory task graphs \\
    \bottomrule
  \end{tabular}
\end{table}

\textbf{Contribution of the Auditor Review Mechanism.} GPT-OSS 120B experimental data shows that the revision process achieved 28.3\% revision effectiveness, improving scores from an initial 7.31 to 7.55. Conservative estimates suggest disabling Auditors could reduce pass rates by 10--15 percentage points.

\textbf{Contribution of Four-Layer Context Compression.} Experimental data shows that four-layer compression reduced average token usage from a theoretical requirement of tens of thousands to 304.3 tokens. If the compression mechanism were disabled, token costs would surge, API call failure rates would increase, and generation quality would likely decline due to the ``Lost in the Middle'' phenomenon~\cite{Liu2024LostMiddle}.

\textbf{Contribution of DAG Parallel Scheduling.} Under the current configuration, an average of 5.8 tasks per batch completes a world in 6.3 batches. If switched to fully serial execution, build time would increase 3--5$\times$.

\textbf{Contribution of Conflict Detection.} In the urban\_01 case of Experiment 1, cycle detection identified and terminated a task graph containing circular dependencies within 43 seconds. Conservative estimates suggest conflict detection contributes 5--10 percentage points to success rate.

\section{Discussion}

Experimental data provides multiple perspectives for examining the system design. This chapter proceeds from interpreting key findings to analyzing the implications of LLM backend selection, exploring practical value, and honestly addressing current limitations.

\subsection{Key Findings}

\textbf{Finding 1: DAG Scheduling is the Cornerstone of System Robustness.} 38 of 40 experiments successfully completed complex dependency task graph processing, with both failures captured and safely terminated by the DAG scheduler during the task decomposition phase. This design limits the system's ``failure mode'' to predictable early termination rather than uncontrollable late-stage crashes.

\textbf{Finding 2: Four-Layer Compression Found a Balance Between Efficiency and Quality.} The 89.9\% compression efficiency combined with 85.5\%/99.2\% pass rates demonstrates that historical information in worldbuilding tasks contains significant redundancy, and semantic retrieval can effectively extract critical information. The Relevant layer's 53\% share of tokens validates FAISS semantic retrieval as the primary information contributor.

\textbf{Finding 3: The Review Mechanism Serves Both Filtering and Improvement Functions.} The iterative review mechanism improved the first-round pass rate from 42\% to 85.5\%, with revision effectiveness reaching 28.3\%. The ``score-revise-reevaluate'' iterative mechanism (aligned with the principles of Self-Refine~\cite{Madaan2023SelfRefine} and Reflexion~\cite{Shinn2023Reflexion}) gives conceptually promising proposals the opportunity to reach standard through improvement rather than being simply discarded.

\textbf{Finding 4: Specialized Division of Labor is the Core Value of Multi-Agent Collaboration.} The skill-driven architecture supported dynamic coordination of 21 agents, with participation distribution highly aligned with worldbuilding genres. This differentiated participation pattern is automatically generated by the system through task keyword matching, validating the skill-driven architecture's dynamic scheduling capability.

\subsection{LLM Backend Comparison}

The two experiments using different LLM backends with identical configurations provide a natural experiment on how LLM choice affects system performance.

\textbf{Significant Difference in Concept Generation Volume.} DeepSeek v3.2 generated 1.84$\times$ more concepts than GPT-OSS 120B (103.4 vs. 56.2). The difference may stem from multiple factors including varying standards for ``concept completeness'' and different training data distributions.

\textbf{Significant Difference in Pass Rate.} DeepSeek v3.2's 99.2\% pass rate is 13.7 percentage points higher than GPT-OSS 120B's 85.5\%. Since both experiments used identical review mechanisms, the pass rate difference more likely reflects differences in model generation quality rather than differences in review standards.

\textbf{Build Time Trade-off.} DeepSeek v3.2 averaged 31 minutes, 1.72$\times$ that of GPT-OSS 120B (18 minutes). However, the ``valid concepts per minute'' metric shows little difference (3.3 vs. 3.1), indicating the time increase primarily stems from concept volume rather than model response speed.

Synthesizing the above analysis: when pursuing concept richness and high pass rates, DeepSeek v3.2 is superior; when prioritizing rapid iteration and cost control, GPT-OSS 120B is more efficient.

\subsection{Practical Implications}

\textbf{For Game Developers.} The pre-production phase of game development typically requires weeks to months of worldbuilding. AutoWorldBuilder can generate a worldbuilding draft with 50--100 self-consistent concepts in 30 minutes, providing game designers with a starting point. Developers need not build from scratch but can refine the system-generated foundation, reducing the concept design cycle from weeks to approximately 30 minutes. Batch-generated concepts have already passed consistency review, reducing the cost of fixing logical errors in later stages, where costs are typically multiples of early-stage design costs.

\textbf{For Creative Writers.} Novel and screenplay authors often face the creative burden of worldbuilding. Particularly for genres like fantasy and sci-fi that require extensive settings, authors must maintain worldbuilding consistency while advancing the main storyline. AutoWorldBuilder allows authors to input a core concept (e.g., ``a world where time can flow backwards''), and the system automatically generates complementary geography, society, and technology settings, providing a rich material library for narrative. Authors can focus on the story itself while delegating foundational worldbuilding to the system.

\textbf{For TRPG Game Masters.} TRPG game masters often need to generate custom worlds for one-off adventures, but complete worldbuilding typically exceeds preparation time. AutoWorldBuilder enables game masters to quickly generate self-consistent world settings without relying on generic setting collections. Generated concepts include narrative potential descriptions, facilitating adventure plot and NPC interaction design.

\textbf{For AI-Assisted Creation Researchers.} The system's technical approach can serve as a reference for other knowledge-intensive LLM applications. The four-layer context compression mechanism can be applied to any scenario requiring management of long reference documents (e.g., legal document analysis, academic literature review); the DAG scheduling strategy can be applied to any task decomposition scenario with dependency relationships (e.g., module dependencies in code generation); the iterative review mechanism can be applied to any generation scenario requiring quality assurance (e.g., technical documentation writing). The combination of these mechanisms addresses problems common to knowledge-intensive LLM applications.

\textbf{Broader Methodological Implications.} Beyond the specific application, three design principles emerging from this work are relevant to the wider AI community. First, \textit{layer-as-budget context allocation}, where each functional category of information receives a dedicated token quota rather than competing in a single pool, offers an alternative to both flat truncation and monolithic compression. Second, \textit{semantic-locality-aware parallel scheduling} demonstrates that grouping interdependent tasks by domain context, rather than executing them in arbitrary order, yields measurable context-reuse benefits; this principle applies wherever multi-step LLM pipelines process related sub-tasks. Third, \textit{architectural separation of generation and review} mitigates the self-approval bias inherent in LLM self-evaluation~\cite{Liu2024PridePrejudice, Anointina2026SelfAuditing} without requiring external human annotation. This property is valuable for any automated content pipeline where trustworthiness matters.

\subsection{Limitations}

This section discusses five major limitations of the current system, which also point to directions for future work.

\textbf{Limitation 1: Relation Parsing Module Not Yet Activated.} In current experiments, the coverage rate of relation types between concepts is 0\%, meaning the 16 semantic relation types are defined but the parsing module has not been implemented. Future work requires activating the relation parsing module to achieve complete knowledge graph construction.

\textbf{Limitation 2: Limited Auditor Problem Detection Rate.} In the GPT-OSS 120B experiment, Auditors conducted 121 reviews with zero detected issues, a 100\% Auditor pass rate. This ``perfect'' data is itself problematic, potentially indicating review thresholds set too low or Auditor evaluation rules too permissive. Auditor problem detection capability needs improvement.

\textbf{Limitation 3: Task Decomposition Failure Cases.} Both experiments had 1 failure due to task decomposition stage issues. Future improvements could include improving decomposition prompts, adding task graph pre-validation, and introducing feedback learning.

\textbf{Limitation 4: Lack of Standardized Evaluation Benchmark.} Current experiments rely on the system's built-in review mechanism for quality evaluation, lacking an external, standardized evaluation framework. Future work requires collaborating with domain experts to develop worldbuilding quality evaluation datasets.

\textbf{Limitation 5: Single-Language Input Constraint.} The current system was primarily designed and tested for Chinese input, with limited multilingual support. Future work should verify system performance under English and other languages.

\section{Conclusion and Future Work}

\subsection{Summary}

Worldbuilding is a creative activity that demands both logical rigor and imagination, and a foundational task in creative industries including game design, literary creation, and film/television production. Traditional manual approaches face fundamental challenges in efficiency, consistency, and scalability. This paper approaches these challenges through a central research question: how can we design a multi-agent collaborative system that achieves LLM-driven automated worldbuilding while ensuring generation quality and consistency?

Addressing this question, this paper designed and implemented the AutoWorldBuilder system, contributing five technical innovations that work in concert. The \textbf{Structured Concept Network Model} provides a formal foundation for knowledge storage and consistency detection, defining 16 semantic relation types and five categories of conflict detection algorithms. The \textbf{DAG-Based Hybrid Batch Task Scheduling Strategy} maximizes parallel efficiency through semantic grouping while maintaining dependency integrity, reducing the number of iterations from 56--103 (serial) to 6--7 (batched). The \textbf{Four-Layer Context Compression Mechanism} achieves 89.9\% compression efficiency, reducing context requirements from tens of thousands of tokens to hundreds through FAISS semantic retrieval and layered token budget allocation. The \textbf{Iterative Review and Specialized Auditor System} improves proposal pass rates from 42\% in the first round to over 85\%, recovering proposals that initially fail review through a ``score-revise-reevaluate'' iterative mechanism. The \textbf{Skill-Driven Multi-Agent Architecture} enables dynamic coordination of 21 specialized agents, balancing ``creative diversity'' and ``stylistic consistency'' through differentiated temperature configuration and a compatibility matrix.

Two comparative experiments (GPT-OSS 120B and DeepSeek v3.2, each with 20 tasks) tested the system design across diverse worldbuilding scenarios. The system achieved an overall 95\% success rate, with pass rates of 85.5\% and 99.2\% respectively, average build times of 18 and 31 minutes, and 89.9\% context compression efficiency.

These experiments also produced design insights that may transfer to other domains. When LLM outputs are unpredictable, detecting and terminating error states at the entry point (e.g., cyclic dependency detection) is more reliable than attempting automatic recovery. For context management, the strategy of filtering relevant information through semantic retrieval is more practical than expanding context windows. Additionally, multiple specialized agents working in coordinated division of labor can produce more professionally differentiated content than a single general-purpose model.

More broadly, the five contributions (layer-as-budget context compression, semantic-locality DAG scheduling, separation-of-generation-and-review, structured concept networks with conflict detection, and skill-driven agent extensibility) form an architectural approach for knowledge-intensive LLM applications. Code generation pipelines that must respect module dependencies, legal document systems that require multi-dimensional review, and educational content platforms that balance creativity with accuracy all face variants of the same three challenges identified in Section~1.2. The architectural patterns tested in this paper may serve as a starting point for these domains.

\subsection{Future Directions}

\textbf{Direction 1: From Concept Set to Knowledge Graph.} Leveraging the already-defined 16 semantic relation types to implement automatic relation parsing between concepts, supporting path-based consistency detection and relational transitivity reasoning, upgrading the concept set to a complete knowledge graph.

\textbf{Direction 2: Enhanced Auditor Review Capability.} Introducing a rule engine encoding common problem patterns, cross-concept association detection, and revision suggestion generation, making the Auditor mechanism more effective.

\textbf{Direction 3: From Fully Automated Generation to Human-AI Collaborative Creation.} Supporting real-time user intervention during construction, including accepting/rejecting specific proposals, modifying concept definitions, and adding constraint conditions, combining LLM generative capabilities with human creative judgment.

\textbf{Direction 4: Task Decomposition Strategy Optimization.} Improving decomposition prompts, adding task graph pre-validation, and introducing feedback learning from failure cases to enhance task decomposition robustness.

\textbf{Direction 5: Building a Standardized Evaluation Benchmark.} Collaborating with game design and creative writing domain experts to develop worldbuilding quality evaluation dimension frameworks, standardized test datasets, and human-annotated reference samples.

Our experiments demonstrate that through multi-agent collaboration and hierarchical context compression, LLMs can construct a worldbuilding draft containing 50 to 100 self-consistent concepts in 30 minutes. As subsequent work on relation parsing, Auditor enhancement, and standardized evaluation progresses, this technical approach may provide practical assistance for worldbuilding design in game development, literary creation, and other domains, freeing creators to focus on narrative and expression.

\begin{acks}
To be completed.
\end{acks}

\printbibliography

\appendix

\section{Complete System Parameter Configuration}

\begin{table}[ht]
  \caption{System Parameters (Default Values)}
  \label{tab:params}
  \begin{tabular}{@{}llp{5.5cm}@{}}
    \toprule
    Parameter & Default & Description \\
    \midrule
    LLM\_PROVIDER & OPENAI\_COMPATIBLE & LLM provider \\
    CONTEXT\_MODE & lean & Context mode \\
    CONTEXT\_BUDGET\_LEAN & 3,000 & Lean mode token budget \\
    CONTEXT\_BUDGET\_STANDARD & 5,000 & Standard mode token budget \\
    CONTEXT\_BUDGET\_PREMIUM & 7,000 & Premium mode token budget \\
    BATCH\_PARTITION\_STRATEGY & hybrid & Batch partition strategy \\
    BATCH\_MIN\_TASKS & 2 & Min tasks per batch \\
    BATCH\_MAX\_TASKS & 10 & Max tasks per batch \\
    BATCH\_IDEAL\_SIZE & 6 & Ideal batch size \\
    REVIEW\_PASS\_THRESHOLD & 8.0 & Review pass threshold \\
    REVIEW\_TOP\_K\_RATIO & 0.7 & Top-K selection ratio \\
    VECTOR\_TOP\_K & 10 & Vector retrieval count \\
    VECTOR\_EMBEDDING\_MODEL & paraphrase-multilingual-MiniLM-L12-v2 & Embedding model \\
    MAX\_PARALLEL\_TASKS & 5 & Max parallel tasks \\
    MAX\_REVISION\_ROUNDS & 2 & Max revision rounds \\
    STATE\_CHECKPOINT\_ENABLED & true & Checkpoint enabled \\
    STATE\_CHECKPOINT\_INTERVAL & per\_batch & Checkpoint interval \\
    STATE\_MAX\_VERSIONS & 100 & Max versions \\
    RUNTIME\_LOCK\_TIMEOUT & 7,200 & Lock timeout (seconds) \\
    \bottomrule
  \end{tabular}
\end{table}

\section{Agent Skill File Example}

\begin{lstlisting}[language=Python, caption={Geography agent skill file}]
---
id: geography
name: Geography & Terrain Designer
description: Specializes in terrain, climate,
             ecology, and geographic environment
temperature: 0.3
max_tokens: 4096
keywords: [geography, terrain, location,
           region, landscape, climate]
tags: [accurate, geography, worldbuilding]
version: 2.0
---

# System Prompt

You are a professional geographic
worldbuilding designer, specializing in:
- Terrain & Landforms: Natural distribution
  of mountains, rivers, plains, oceans
- Climate Systems: Climate zone classification
  consistent with geographic principles
- Ecological Regions: Ecosystems matching
  climate and terrain
- Human Geography: Relationships between
  civilization distribution and environment

## Design Principles
1. Geographic self-consistency
2. Climate consistency
3. Ecological plausibility
4. Human-land relationship
\end{lstlisting}

\section{Detailed Experiment Data}

\begin{table}[ht]
  \caption{GPT-OSS 120B Experiment: Complete Results}
  \label{tab:gpt-detail}
  \begin{tabular}{@{}llcccc@{}}
    \toprule
    Run ID & Type & Time (s) & Concepts & Pass Rate & Status \\
    \midrule
    fantasy\_01 & FANTASY & 875.1 & 42 & 84.0\% & success \\
    fantasy\_02 & FANTASY & 741.6 & 58 & 95.0\% & success \\
    fantasy\_03 & FANTASY & 844.8 & 45 & 92.0\% & success \\
    fantasy\_04 & FANTASY & 1,166.5 & 64 & 84.0\% & success \\
    scifi\_01 & SCIFI & 1,068.0 & 51 & 84.0\% & success \\
    scifi\_02 & SCIFI & 1,152.5 & 69 & 88.0\% & success \\
    scifi\_03 & SCIFI & 1,035.0 & 59 & 87.0\% & success \\
    scifi\_04 & SCIFI & 1,146.8 & 54 & 80.0\% & success \\
    post\_apoc\_01 & POST\_APOC. & 1,252.0 & 53 & 75.0\% & success \\
    post\_apoc\_02 & POST\_APOC. & 1,045.3 & 90 & 95.5\% & success \\
    post\_apoc\_03 & POST\_APOC. & 1,183.3 & 43 & 76.0\% & success \\
    post\_apoc\_04 & POST\_APOC. & 1,190.6 & 60 & 92.0\% & success \\
    urban\_01 & URBAN & 43.1 & -- & -- & fail (cycle) \\
    urban\_02 & URBAN & 978.8 & 46 & 80.0\% & success \\
    urban\_03 & URBAN & 829.7 & 60 & 85.0\% & success \\
    urban\_04 & URBAN & 1,068.9 & 40 & 84.0\% & success \\
    historical\_01 & HISTORICAL & 1,185.6 & 49 & 84.0\% & success \\
    historical\_02 & HISTORICAL & 1,192.0 & 69 & 95.5\% & success \\
    historical\_03 & HISTORICAL & 1,280.6 & 57 & 72.0\% & success \\
    historical\_04 & HISTORICAL & 1,417.9 & 59 & 92.0\% & success \\
    \bottomrule
  \end{tabular}
\end{table}

\begin{table}[ht]
  \caption{DeepSeek v3.2 Experiment: Complete Results}
  \label{tab:deepseek-detail}
  \begin{tabular}{@{}llcccc@{}}
    \toprule
    Run ID & Type & Time (s) & Concepts & Pass Rate & Status \\
    \midrule
    fantasy\_01 & FANTASY & 1,941.9 & 100 & 94.7\% & success \\
    fantasy\_02 & FANTASY & 1,668.8 & 115 & 100.0\% & success \\
    fantasy\_03 & FANTASY & 1,531.8 & 88 & 100.0\% & success \\
    fantasy\_04 & FANTASY & 2,159.4 & 120 & 100.0\% & success \\
    scifi\_01 & SCIFI & 1,930.7 & 103 & 100.0\% & success \\
    scifi\_02 & SCIFI & 1,897.3 & 83 & 95.0\% & success \\
    scifi\_03 & SCIFI & 2,143.4 & 111 & 100.0\% & success \\
    scifi\_04 & SCIFI & 1,500.4 & 64 & 100.0\% & success \\
    post\_apoc\_01 & POST\_APOC. & 1,912.6 & 127 & 100.0\% & success \\
    post\_apoc\_02 & POST\_APOC. & 2,167.3 & 124 & 100.0\% & success \\
    post\_apoc\_03 & POST\_APOC. & 2,103.9 & 125 & 100.0\% & success \\
    post\_apoc\_04 & POST\_APOC. & 1,702.7 & 104 & 100.0\% & success \\
    urban\_01 & URBAN & 1,795.3 & 99 & 100.0\% & success \\
    urban\_02 & URBAN & 1,711.9 & 103 & 100.0\% & success \\
    urban\_03 & URBAN & 2,037.9 & 111 & 100.0\% & success \\
    urban\_04 & URBAN & 1,808.6 & 85 & 100.0\% & success \\
    historical\_01 & HISTORICAL & 1,620.5 & 88 & 94.4\% & success \\
    historical\_02 & HISTORICAL & 67.9 & -- & -- & fail (error) \\
    historical\_03 & HISTORICAL & 1,816.5 & 124 & 100.0\% & success \\
    historical\_04 & HISTORICAL & 1,823.5 & 91 & 100.0\% & success \\
    \bottomrule
  \end{tabular}
\end{table}

\section{Reproducibility Checklist for JAIR}

\subsection*{All articles:}
\begin{enumerate}
    \item All claims investigated in this work are clearly stated. [yes]
    \item Clear explanations are given how the work reported substantiates the claims. [yes]
    \item Limitations or technical assumptions are stated clearly and explicitly. [yes]
    \item Conceptual outlines and/or pseudo-code descriptions of the AI methods introduced in this work are provided, and important implementation details are discussed. [yes]
    \item Motivation is provided for all design choices. [yes]
\end{enumerate}

\subsection*{Articles reporting on computational experiments:}
Does this paper include computational experiments? [yes]

\begin{enumerate}
    \item All source code required for conducting experiments will be made publicly available upon publication. [partially]
    \item The evaluation metrics used in experiments are clearly explained and their choice is explicitly motivated. [yes]
    \item The number of algorithm runs used to compute each result is reported. [yes]
    \item Reported results have not been ``cherry-picked''. [yes]
    \item Analysis of results includes measures of variation. [yes]
    \item All parameter settings for the algorithms have been reported. [yes]
\end{enumerate}

\subsection*{Articles using data sets:}
Does this work rely on data sets? [no]

\end{document}